\documentclass[10pt,twocolumn,letterpaper]{article}

\usepackage[pagenumbers]{cvpr}   

\usepackage[dpi=300,authorsoneline]{latexml}
\iflatexml
\else
\usepackage{xeCJK}
\setCJKsansfont{FandolHei-Regular.otf}[BoldFont=FandolHei-Bold.otf]
\setCJKmonofont{FandolFang-Regular.otf}
\fi
\usepackage{multirow,array,tabularx,colortbl}
\definecolor{retacgray}{gray}{0.91}
\newcolumntype{C}{>{\centering\arraybackslash}X}
\usepackage{placeins,float,afterpage}
\newcommand{\supphead}[1]{\begin{tabular}[c]{@{}c@{}}#1\end{tabular}}

\definecolor{cvprblue}{rgb}{0.21,0.49,0.74}
\usepackage[pagebackref,breaklinks,colorlinks,allcolors=cvprblue]{hyperref}

\def\confName{CVPR}
\def\confYear{2027}

\title{MXI-ReTac: When to Intervene and How to Correct in Contact-Rich Visuotactile Manipulation}
\iflatexml
\author{\begin{tabular}{c}
Wen Xiao$^{1,2}$, Shihao Zhou$^{1}$, Jian Qiao$^{1}$, Wenyuan Sheng$^{1}$, Feng Yan$^{1}$,\\
Jialong Peng$^{1}$, Yuxin Liu$^{2}$, Haozheng Wang$^{1}$, Lin Ma$^{1,*}$, Chao Yang$^{1}$\\[4pt]
$^{1}$1011.ai\qquad $^{2}$Central South University\\[4pt]
$^{*}$Corresponding author
\end{tabular}}
\else
\author{
Wen Xiao$^{1,2}$\quad Shihao Zhou$^{1}$\quad Jian Qiao$^{1}$\quad Wenyuan Sheng$^{1}$\quad Feng Yan$^{1}$\\
Jialong Peng$^{1}$\quad Yuxin Liu$^{2}$\quad Haozheng Wang$^{1}$\quad Lin Ma$^{1}$\thanks{Corresponding author.}\quad Chao Yang$^{1}$\\[4pt]
{\normalsize $^{1}$1011.ai\quad $^{2}$Central South University}
}
\fi

\begin{document}
\raggedbottom
\maketitle

\begin{abstract}
Tactile signals have been widely studied for improving vision-language-action (VLA) policies. However, when and how to apply tactile corrections during action-chunk execution remain underexplored. We first build \textbf{Tactile Simple Baseline} from existing designs as a reference policy with both tactile history and within-chunk residual feedback. To address how to correct, we extract correction targets from disagreements between overlapping successive plans without additional data collection. From these targets, \textbf{Replanning-Disagreement Distillation (RD)} trains a lightweight head for execution-time residual correction. To address when to correct, we integrate this head into \textbf{ReTac}, a reliability-aware visuotactile policy. During planning, its \textbf{Action Validity Expert (AVE)} weights visual and tactile attention by observation reliability. During execution, AVE chooses among keeping the plan, applying residual corrections, and replanning early, while continuously adjusting residual strength. Across nine simulated tasks, ReTac achieves mean success rates of $87.8\%$ on six UniVTAC tasks and $64.2\%$ on three OpenViTac tasks. In zero-shot cross-embodiment transfer without target fine-tuning, it reaches $66.0\%$ across arms and $77.7\%$ across grippers. Finally, it averages $63.8\%$ across four real-world UMI tasks.
\end{abstract}

\begin{figure}[!t]
\centering
\iflatexml
\includegraphics[width=\columnwidth]{retac-policy-overview-html.svg}
\else
\includegraphics[width=\columnwidth]{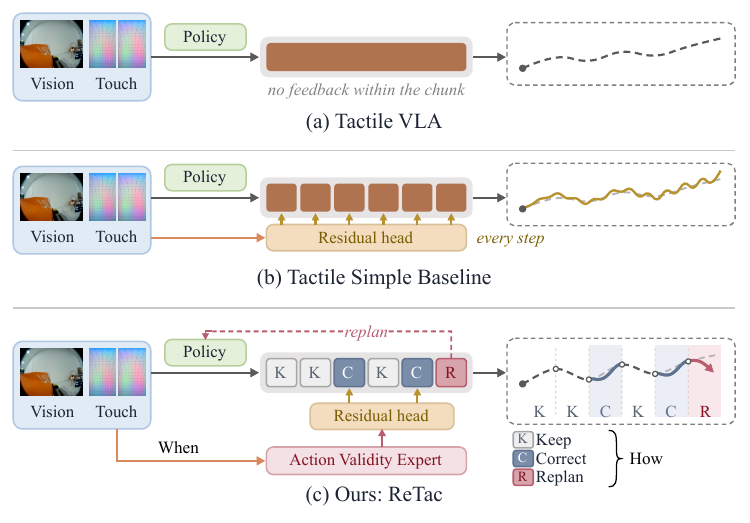}
\fi
\caption{From tactile feedback to selective intervention. (a) Tactile VLA generates reference action chunks. (b) Tactile Simple Baseline adds residual feedback, which can still produce harmful corrections. (c) ReTac jointly regulates modality weights, intervention timing, and residual strength through AVE and a replanning-disagreement-distilled (RD) head.}
\label{fig:timeline}
\end{figure}

\section{Introduction}

Contact-rich manipulation depends heavily on local tactile feedback. External cameras provide global target poses and scene geometry, whereas fingertip optical touch reveals deformation, microslip, and resistance at occluded interfaces during insertion, grasping, and extraction~\cite{yuan2017gelsight,lambeta2020digit,taylor2021gelslim}. Vision-language-action (VLA) models~\cite{kim2024openvla,black2024pi0,pi2025pi05} typically fuse multimodal observations to generate multi-step action chunks executed open-loop. Yet contact establishment, slip, and jamming often occur between planning calls and cannot promptly affect ongoing actions. Tactile signals have been widely studied for improving VLA policies, but when and how to apply tactile corrections during chunk execution remain underexplored.

Existing methods struggle to address both questions. At planning time, SimpleTouch~\cite{yang2026simpletouch}, FTP-1~\cite{yuan2026ftp1}, and OpenViTac~\cite{wu2026openvitac} introduce tactile representations or auxiliary tactile prediction, while VT-MUSE~\cite{vtmuse} and FM-VLA~\cite{fmvla} encode visual--tactile or force histories. These methods improve reference trajectories but retain open-loop execution within each chunk. At execution time, T-Rex~\cite{trex} invokes a tactile expert on a fixed schedule to continue flow-matching denoising; ViTaR~\cite{wang2026vitar}, VLA-Touch~\cite{bi2025vlatouch}, and TacForcing~\cite{zhou2026tacforcing} refine slow-policy outputs through residuals, lightweight adaptation, or streaming generation. ResiP~\cite{ankile2024resip}, RL Token~\cite{xu2026rltoken}, Real-Time EXPO-FT~\cite{expoft}, and OmniVTA~\cite{omnivta} instead incorporate reinforcement learning or recovery-trajectory supervision. Such approaches rely on preset schedules or fixed correction rates and gains, or require online interaction and manually collected recovery data. Contact-based gating, as in TacVLA~\cite{tacvla}, switches tactile pathways, but physical contact does not necessarily imply that an action needs correction.

To ensure that improvements are evaluated against a reference equipped with both planning- and execution-time tactile feedback, we integrate effective existing designs into \textbf{Tactile Simple Baseline (TSB)}. Building on SimpleTouch's pretrained tactile representations and future tactile prediction, TSB adds cross-step tactile memory during planning and a ResiP residual head during execution. With the main policy frozen, the simulation ResiP head is trained with PPO under a matched episode budget and provides within-chunk feedback at a fixed interval and gain. TSB thus establishes a common starting point for evaluating replanning-disagreement distillation and adaptive gating.

To address how to correct, our key observation is that successive predictions for an overlapping time window directly capture how the same policy revises its reference trajectory after receiving new feedback. These disagreements provide supervision using only existing demonstrations, without online interaction or additional recovery data. We introduce \textbf{Replanning-Disagreement Distillation (RD)} to distill these revisions into a lightweight tactile correction head, which predicts local action residuals from current touch and proprioception. A tactile-encoding-plus-correction call takes $8.32\,\text{ms}$, or $12.73\,\text{ms}$ when including AVE decision-making, compared with $120.35\,\text{ms}$ per main-policy planning call, enabling low-latency feedback between slower planning updates.

To address when to correct, experiments reveal conflicting intervention requirements across tasks, stages, and embodiments. Reducing the correction interval from five to two steps raises key-extraction success from $85.0\%$ to $100.0\%$, but lowers tube-insertion success from $90.0\%$ to $80.0\%$. In physical manipulation, cylinder grasping benefits from timely adjustments against slip, whereas additional disturbance after securing a potato chip can crush it. Changes in arm kinematics and gripper structure can also make source-embodiment corrections inappropriate. Fixed gains or contact-triggered gates cannot accommodate all these conditions. We therefore introduce an Action Validity Expert (AVE) for two-level decision-making, which combines with RD to form \textbf{ReTac} (Fig.~\ref{fig:timeline}). During planning, AVE weights visual and tactile attention according to observation reliability and tactile continuity. During execution, it compares past tactile predictions with current measurements and combines them with motion state to estimate correction utility, selecting among KEEP, CORRECT, and REPLAN while continuously scaling residual strength.

Experiments validate these designs across multiple settings. ReTac achieves mean success rates of $87.8\%$ on six UniVTAC tasks~\cite{chen2026univtac} and $64.2\%$ on three OpenViTac tasks. In zero-shot cross-embodiment transfer without target fine-tuning, it reaches $66.0\%$ from Franka to UR5e and $77.7\%$ from the Panda gripper to the Allegro hand via a thumb--index pinch mapping. It averages $63.8\%$ across cylinder grasping, potato-chip grasping, cap twisting, and medication bagging on UMI. Across 60 trials with matched scene seeds, ReTac achieves $85.0\%$ success, exceeding vision-only GPT-6 Astra~\cite{yang2026astra} at $76.7\%$ while remaining below Astra with vision and touch at $95.0\%$. Local planning and tactile correction take $120.35\,\text{ms}$ and $8.32\,\text{ms}$ per call, respectively, compared with $21.13\,\text{s}$ per Astra API request.

Our contributions are threefold.
\iflatexml
\begin{enumerate}[label=\textbullet]
    \item We propose \textbf{Replanning-Disagreement Distillation (RD)}. It distills the planner's revisions to overlapping trajectories into a lightweight residual head. The head enables fast within-chunk correction. Training uses only offline demonstrations, without additional interaction or recovery data.
    \item We introduce an \textbf{Action Validity Expert (AVE)} for planning-time modality weighting and execution-time selection among keeping, correcting, and replanning with adaptive residual strength. Together, AVE and RD form \textbf{ReTac}, coordinating local correction with global trajectory stability.
    \item We construct \textbf{Tactile Simple Baseline} with tactile history and residual feedback. Extensive benchmark and real-world evaluations demonstrate ReTac's improved contact-rich manipulation performance and zero-shot cross-embodiment adaptability.
\end{enumerate}
\else
\begin{itemize}
    \item We propose \textbf{Replanning-Disagreement Distillation (RD)}. It distills the planner's revisions to overlapping trajectories into a lightweight residual head. The head enables fast within-chunk correction. Training uses only offline demonstrations, without additional interaction or recovery data.
    \item We introduce an \textbf{Action Validity Expert (AVE)} for planning-time modality weighting and execution-time selection among keeping, correcting, and replanning with adaptive residual strength. Together, AVE and RD form \textbf{ReTac}, coordinating local correction with global trajectory stability.
    \item We construct \textbf{Tactile Simple Baseline} with tactile history and residual feedback. Extensive benchmark and real-world evaluations demonstrate ReTac's improved contact-rich manipulation performance and zero-shot cross-embodiment adaptability.
\end{itemize}
\fi

\section{Related Work}

\subsection{Visuotactile VLA Policies}
Building on generalist robot learning~\cite{brohan2022rt1,brohan2023rt2,octo2024}, VLA models such as OpenVLA~\cite{kim2024openvla}, $\pi_0$~\cite{black2024pi0}, and $\pi_{0.5}$~\cite{pi2025pi05} generate multi-step action chunks through multimodal pretraining. Because vision alone misses deformation, slip, and jamming at occluded contacts, prior work combines vision and touch for grasping, insertion, and 3D manipulation~\cite{calandra2018feeling,lee2019sense,huang2024vitac}. Representation and benchmark efforts include Touch and Go~\cite{yang2022touchgo}, Sparsh~\cite{higuera2024sparsh}, UniTouch~\cite{yang2024unitouch}, AnyTouch~\cite{feng2025anytouch}, UniVTAC~\cite{chen2026univtac}, ManiFeel~\cite{luu2025manifeel}, and OpenViTac~\cite{wu2026openvitac}. Recent policies incorporate pretrained tactile features, future tactile prediction, equivariant diffusion, or contact-history and world modeling~\cite{yang2026simpletouch,yuan2026ftp1,wong2026vista,fmvla,vtmuse,robopack,vtwm}, while GPT-6 Astra~\cite{yang2026astra} reasons over vision and touch via second-scale cloud requests. These planning-side designs improve reference trajectories but retain open-loop execution within each chunk. Our TSB extends SimpleTouch with tactile history and ResiP correction to provide feedback in both planning and execution.

\subsection{Execution-Time Tactile Correction}
Residual policy learning~\cite{silver2018residual,johannink2019residual} refines base controllers using three main supervision sources: offline demonstrations for fast refinement or streaming updates~\cite{trex,wang2026vitar,bi2025vlatouch,zhou2026tacforcing}, online reinforcement learning~\cite{ankile2024resip,xu2026rltoken,expoft}, and manually collected recovery segments~\cite{omnivta}. In simulation, TSB follows ResiP with online PPO after freezing the planner, whereas ReTac replaces this head with RD trained solely on existing offline demonstrations.

General action-chunk execution methods also mitigate open-loop lag and inference delay through receding-horizon replanning~\cite{chi2023diffusion}, temporal ensembling~\cite{zhao2023act}, closed-loop resampling~\cite{liu2024bid}, and asynchronous flow-matching inference~\cite{black2025rtc,vlash,futurertc}. While FutureRTC~\cite{futurertc} uses policy consistency for latency compensation, RD distills successive replanning revisions in action space for tactile-driven within-chunk correction from ordinary demonstrations.

\subsection{Adaptive Gating}
Existing gating methods regulate tactile feedback via contact detection~\cite{tacvla,omnivta}, adaptive tactile or predicted-force attention~\cite{li2026atvla,pfa}, force-prediction deviations~\cite{fawam}, or active belief-space probing~\cite{activebelief}. However, contact or force deviation alone does not guarantee that a local residual is beneficial: unnecessary corrections can disrupt stable grasps or crush fragile objects, whereas large deviations require full replanning. AVE instead weights modalities by observation reliability during planning and predicts correction utility during execution to choose among KEEP, CORRECT, and REPLAN with continuous residual scaling.

\section{Method}

\subsection{Problem Formulation}
\label{sec:problem_formulation}
We consider closed-loop manipulation with visual and tactile feedback, executed in action chunks. At planning time $t_0$, the main policy predicts $H$ reference actions from the instruction, visual observations, tactile history, and proprioceptive state, executing at most $N$ steps before updating the plan. Given feedback within a chunk, we distinguish two questions: \textbf{how to correct}, by predicting a local action residual $\Delta\hat a_t$; and \textbf{when to correct}, by selecting an operation $m_t\in\{\mathrm K,\mathrm C,\mathrm R\}$ and a residual scaling gain $\alpha_t$. Here, $\mathrm K$, $\mathrm C$, and $\mathrm R$ denote keeping the plan, applying a residual correction, and replanning early, respectively.

\begin{figure*}[t]
\centering
\iflatexml
\includegraphics[width=\textwidth]{retac-square-revised-html.svg}
\else
\includegraphics[width=\textwidth]{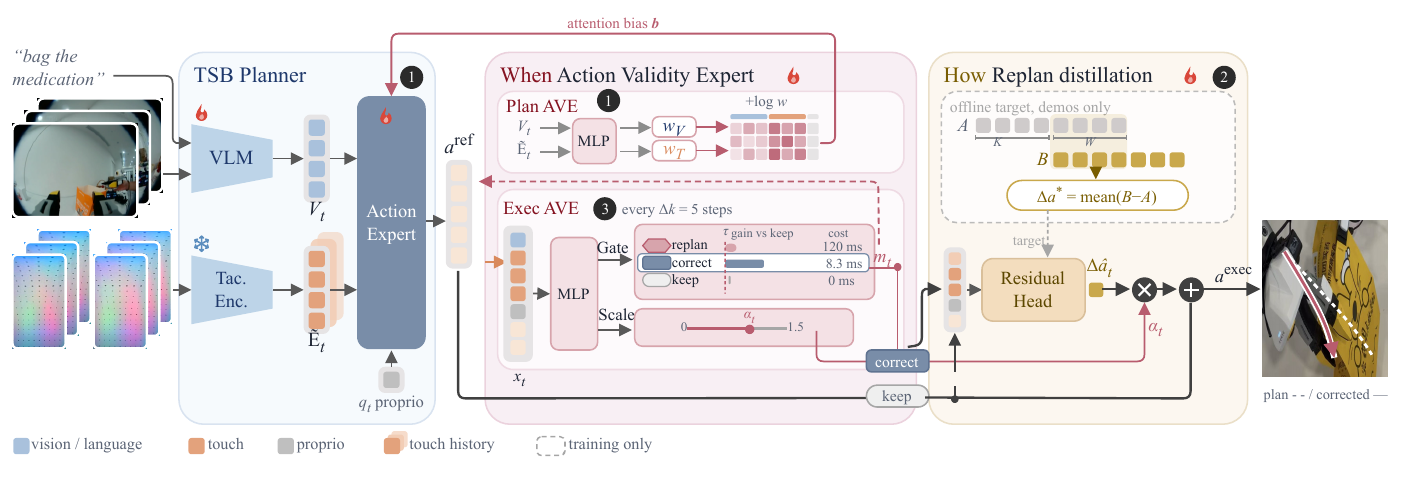}
\fi
\caption{ReTac architecture. The policy combines tactile history fusion, chunk-level planning, replanning-disagreement-distilled (RD) residual correction, and two-level Action Validity Experts, namely Plan AVE and Exec AVE. Numbered badges \textbf{1}--\textbf{3} indicate the three offline training stages; flame and snowflake icons denote trainable and frozen modules, respectively.}
\label{fig:architecture}
\end{figure*}

Figures~\ref{fig:timeline} and~\ref{fig:architecture} illustrate the framework. We present Tactile Simple Baseline in Sec.~\ref{sec:baseline}, replanning-disagreement distillation in Sec.~\ref{sec:replan_distillation}, AVE in Sec.~\ref{sec:retac_ave}, and the training procedure in Sec.~\ref{sec:training}.

\subsection{Tactile Simple Baseline}
\label{sec:baseline}
\label{sec:error_analysis}
\label{sec:seesaw_analysis}
We construct Tactile Simple Baseline (TSB) to equip the comparison method with both tactile history and within-chunk correction. TSB adopts flow-matching action chunks from $\pi_{0.5}$~\cite{pi2025pi05} and the frozen tactile encoder from SimpleTouch~\cite{yang2026simpletouch}. A linear projection produces the current tactile features $E$, which are fused with history through multi-head cross-attention:
\begin{equation}
\widetilde E=\mathrm{LN}\!\left(E+\mathrm{CrossAttn}(E_{\mathrm Q},S_{\mathrm K},S_{\mathrm V})\right).
\label{eq:tac_cross_attn}
\end{equation}
$S$ concatenates uniformly sampled historical features with temporal position embeddings. Multi-head cross-attention retrieves relevant history; its output is added to $E$ and layer-normalized. The main policy concatenates visual features and $\widetilde E$ without modality-specific attention bias, and generates action chunks conditioned on the instruction and proprioceptive state.

For execution, TSB adopts the ResiP residual policy~\cite{ankile2024resip}, which is decoupled from the main planner. In simulation, we freeze the planner and train the residual policy with PPO~\cite{schulman2017ppo}, using a fixed correction interval and unit gain.

\subsection{Replanning-Disagreement Distillation}
\label{sec:replan_distillation}
Replanning incorporates new feedback at the cost of a full policy inference. We supervise a lightweight correction head from disagreements between an earlier plan and time-aligned replanned trajectories over their overlapping window.

Let $A$ be the reference chunk generated at $t_0$ with horizon $H$, and let $\mathcal{T}_{t_0}=\{t_0+5,t_0+10,\dots,t_0+N\}$ denote the five-step correction times. At each $t\in\mathcal{T}_{t_0}$ with offset $g=t-t_0$, let $B^{(t)}$ be the replanned chunk generated from the demonstration observation at step $t$. Along the same trajectory, $A$ and $B^{(t)}$ share the initial flow-matching noise sample to remove independent sampling variance. With overlap length $W=5$, the index satisfies $g+W\le N+W\le H$ within the chunk horizon, yielding the per-window target and loss:
\begin{equation}
\begin{aligned}
\Delta a_t^\star &= \frac{1}{W}\sum_{j=0}^{W-1}\left(B_j^{(t),\mathrm{arm}}-A_{g+j}^{\mathrm{arm}}\right), \\[-1pt]
&\hspace{5em} g = t - t_0,\\
\mathcal L_{\mathrm{RD}} &= \frac{1}{|\mathcal T_{t_0}|}\sum_{t\in\mathcal T_{t_0}}
\mathrm{SmoothL1}\!\left(\Delta\hat a_t,\Delta a_t^\star\right).
\end{aligned}
\label{eq:replan_target}
\end{equation}
Both $\mathrm{SmoothL1}$ inputs are normalized by per-dimension training scales. Each window $t\in\mathcal{T}_{t_0}$ receives its own time-aligned target comparing $B_j^{(t)}$ and $A_{g+j}$ at step $t+j$ in the same joint-target space.

Let $z_t$ denote pooled tactile features. From planning-time and current touch, tactile and proprioceptive changes, and the remaining reference plan, the correction head predicts one residual for the next five-step window:
\[
\Delta\hat a_t=f_\phi\!\left(z_{t_0},z_t,\Delta z_t,\Delta q_t,
\mathrm{Flatten}(A_t^{\mathrm{ref}})\right).
\]
At inference, this residual updates every five steps via $a^{\mathrm{exec}}=a^{\mathrm{ref}}+\alpha_t\Delta\hat a_t$. Standard discrete-gripper tasks such as UniVTAC apply a 7-DoF arm residual and keep reference gripper commands, whereas the fragile potato-chip tasks, OpenViTac \textit{Grasp Chip} and real-world \textit{Grasp a Potato Chip}, extend both RD and ResiP to 8-D for continuous finger-position residuals, as detailed in Supplementary Sec.~E.3. Supervision uses only offline demonstration predictions without interaction, rewards, or recovery labels. Because $A$ and $B^{(t)}$ are evaluated on offline observations rather than closed-loop rollouts, RD distills the planner's trajectory revision direction under new observations rather than closed-loop failure-recovery error, leaving Exec AVE to regulate intervention timing and residual scaling.

\subsection{Action Validity Experts}
\label{sec:retac_ave}
Building on history fusion in Sec.~\ref{sec:baseline} and residual correction in Sec.~\ref{sec:replan_distillation}, AVE determines when to intervene at two levels: Plan AVE identifies which modality to trust during planning, while Exec AVE decides whether and how strongly to deviate from the reference trajectory (Fig.~\ref{fig:architecture}).

\textbf{Plan Action Validity Expert.} Plan AVE combines the current tactile query with visual and tactile reliability to produce modality weights $w_V,w_T$. The two reliability estimates are
\begin{equation}
\begin{aligned}
r_V&=\sigma\!\left(f_V(v,c_{\mathrm{mv}},e_{\mathrm{track}})\right),\\
r_T&=\sigma\!\left(f_T(c_{\mathrm{temp}},\delta_T,e_{\mathrm{pred}})\right).
\end{aligned}
\label{eq:plan_reliability}
\end{equation}
The visual branch uses current visual features $v$, cross-view consistency $c_{\mathrm{mv}}$, and the mean tracking error $e_{\mathrm{track}}$ over the preceding chunk. The tactile branch uses temporal continuity $c_{\mathrm{temp}}$, cross-step feature changes $\delta_T$, and tactile prediction error $e_{\mathrm{pred}}$. For cross-embodiment transfer, tracking error is computed in the aligned end-effector pose space. Complete definitions of these six features and the bounded mapping $(r_V,r_T)\to(w_V,w_T)$ are provided in Supplementary Sec.~E.1.

\iflatexml
\begin{table}[t]
\centering
\caption{UniVTAC success rates (SR, \%) and ranks across six tasks. Ties share a rank. Methods are ranked by mean SR in the last column group. TSB denotes Tactile Simple Baseline with a ResiP head. Throughout the tables, $\dagger$ denotes previously reported results and bold indicates the best scores within each setting.}
\label{tab:single_embodiment}
{\footnotesize
{\footnotesize
\begin{tabular}{@{}l|cc|cc|cc|cc|cc|cc|cc@{}}
\toprule
\multirow{2}{*}{\mbox{Method}} & \multicolumn{2}{c|}{\begin{tabular}[c]{@{}c@{}}\textbf{Insert Hole}\end{tabular}} & \multicolumn{2}{c|}{\begin{tabular}[c]{@{}c@{}}\textbf{Insert Tube}\end{tabular}} & \multicolumn{2}{c|}{\begin{tabular}[c]{@{}c@{}}\textbf{Lift Bottle}\end{tabular}} & \multicolumn{2}{c|}{\begin{tabular}[c]{@{}c@{}}\textbf{Lift Can}\end{tabular}} & \multicolumn{2}{c|}{\begin{tabular}[c]{@{}c@{}}\textbf{Pull Out Key}\end{tabular}} & \multicolumn{2}{c|}{\begin{tabular}[c]{@{}c@{}}\textbf{Put Bottle}\\\textbf{in Shelf}\end{tabular}} & \multicolumn{2}{c}{\begin{tabular}[c]{@{}c@{}}\textbf{Avg.}\end{tabular}} \\
\cmidrule{2-15}
 & $\boldsymbol{\mathsf{SR}\mathord{\uparrow}}$ & $\boldsymbol{\mathsf{Rank}\mathord{\downarrow}}$ & $\boldsymbol{\mathsf{SR}\mathord{\uparrow}}$ & $\boldsymbol{\mathsf{Rank}\mathord{\downarrow}}$ & $\boldsymbol{\mathsf{SR}\mathord{\uparrow}}$ & $\boldsymbol{\mathsf{Rank}\mathord{\downarrow}}$ & $\boldsymbol{\mathsf{SR}\mathord{\uparrow}}$ & $\boldsymbol{\mathsf{Rank}\mathord{\downarrow}}$ & $\boldsymbol{\mathsf{SR}\mathord{\uparrow}}$ & $\boldsymbol{\mathsf{Rank}\mathord{\downarrow}}$ & $\boldsymbol{\mathsf{SR}\mathord{\uparrow}}$ & $\boldsymbol{\mathsf{Rank}\mathord{\downarrow}}$ & $\boldsymbol{\mathsf{SR}\mathord{\uparrow}}$ & $\boldsymbol{\mathsf{Rank}\mathord{\downarrow}}$ \\
\midrule
\midrule
\mbox{$\pi_{0.5}$} & \mbox{27.0} & \mbox{8} & \mbox{37.0} & \mbox{9} & \mbox{91.0} & \mbox{6} & \mbox{80.0} & \mbox{3} & \mbox{42.0} & \mbox{8} & \mbox{36.0} & \mbox{8} & \mbox{52.2} & \mbox{6} \\
\mbox{FTP-1$^\dagger$} & \mbox{64.0} & \mbox{5} & \mbox{79.0} & \mbox{2} & \mbox{97.0} & \mbox{2} & \mbox{65.0} & \mbox{6} & \mbox{48.0} & \mbox{6} & \mbox{47.0} & \mbox{5} & \mbox{66.7} & \mbox{4} \\
\mbox{ManiFeel$^\dagger$} & \mbox{32.5} & \mbox{7} & \mbox{45.0} & \mbox{7} & \mbox{55.0} & \mbox{10} & \mbox{10.0} & \mbox{9} & \mbox{\textbf{95.0}} & \mbox{\textbf{1}} & \mbox{40.0} & \mbox{7} & \mbox{46.3} & \mbox{7} \\
\mbox{RDP$^\dagger$} & \mbox{23.0} & \mbox{9} & \mbox{75.0} & \mbox{4} & \mbox{84.0} & \mbox{8} & \mbox{12.0} & \mbox{8} & \mbox{18.0} & \mbox{10} & \mbox{41.0} & \mbox{6} & \mbox{42.2} & \mbox{8} \\
\mbox{ViTaR$^\dagger$} & \mbox{---} & \mbox{---} & \mbox{69.0} & \mbox{6} & \mbox{88.0} & \mbox{7} & \mbox{44.0} & \mbox{7} & \mbox{55.0} & \mbox{5} & \mbox{35.0} & \mbox{9} & \mbox{---} & \mbox{---} \\
\mbox{VT-MUSE$^\dagger$} & \mbox{68.0} & \mbox{4} & \mbox{---} & \mbox{---} & \mbox{84.0} & \mbox{8} & \mbox{---} & \mbox{---} & \mbox{38.0} & \mbox{9} & \mbox{---} & \mbox{---} & \mbox{---} & \mbox{---} \\
\mbox{$\pi_{0.5}$ + LingBot-Touch Tiny} & \mbox{60.0} & \mbox{6} & \mbox{45.0} & \mbox{7} & \mbox{95.0} & \mbox{3} & \mbox{73.0} & \mbox{4} & \mbox{48.0} & \mbox{6} & \mbox{\textbf{65.0}} & \mbox{\textbf{1}} & \mbox{64.3} & \mbox{5} \\
\mbox{SimpleTouch} & \mbox{98.0} & \mbox{2} & \mbox{78.0} & \mbox{3} & \mbox{94.0} & \mbox{5} & \mbox{\textbf{83.0}} & \mbox{\textbf{1}} & \mbox{59.0} & \mbox{4} & \mbox{52.0} & \mbox{3} & \mbox{77.3} & \mbox{3} \\
\midrule
\mbox{TSB} & \mbox{98.0} & \mbox{2} & \mbox{75.0} & \mbox{4} & \mbox{95.0} & \mbox{3} & \mbox{68.0} & \mbox{5} & \mbox{\textbf{95.0}} & \mbox{\textbf{1}} & \mbox{50.0} & \mbox{4} & \mbox{80.2} & \mbox{2} \\
\rowcolor{retacgray}
\mbox{\textbf{ReTac}} & \mbox{\textbf{100.0}} & \mbox{\textbf{1}} & \mbox{\textbf{90.0}} & \mbox{\textbf{1}} & \mbox{\textbf{100.0}} & \mbox{\textbf{1}} & \mbox{82.0} & \mbox{2} & \mbox{\textbf{95.0}} & \mbox{\textbf{1}} & \mbox{60.0} & \mbox{2} & \mbox{\textbf{87.8}} & \mbox{\textbf{1}} \\
\bottomrule
\end{tabular}
}
}
\end{table}

\begin{table}[t]
\centering
\caption{Cross-embodiment success rates (\%) without target-embodiment fine-tuning.}
\label{tab:cross_embodiment}
{\footnotesize
\begin{tabular}{lcccccccc}
\toprule
Transfer & Method & Insert Hole & Insert Tube & Lift Bottle & Lift Can & Pull Out Key & \shortstack{Put Bottle\\in Shelf} & Avg. \\
\midrule
\multirow{4}{*}{Franka $\to$ UR5e} & SimpleTouch & \textbf{98.0} & 80.0 & 23.0 & 15.0 & 20.0 & 43.0 & 46.5 \\
 & Fuse & 93.0 & 75.0 & 30.0 & 38.0 & 63.0 & 33.0 & 55.3 \\
 & TSB & \textbf{98.0} & 70.0 & 10.0 & 35.0 & 50.0 & 43.0 & 51.0 \\
 & ReTac & \textbf{98.0} & \textbf{83.0} & \textbf{48.0} & \textbf{44.0} & \textbf{73.0} & \textbf{50.0} & \textbf{66.0} \\
\midrule
\multirow{4}{*}{\shortstack[l]{Panda gripper\\$\to$ Allegro hand}} & SimpleTouch & \textbf{93.0} & \textbf{80.0} & 85.0 & 63.0 & 23.0 & 38.0 & 63.7 \\
 & Fuse & 80.0 & 78.0 & 93.0 & 75.0 & 23.0 & 63.0 & 68.7 \\
 & TSB & \textbf{93.0} & \textbf{80.0} & 70.0 & 75.0 & \textbf{53.0} & \textbf{70.0} & 73.5 \\
 & ReTac & 85.0 & \textbf{80.0} & \textbf{98.0} & \textbf{83.0} & 50.0 & \textbf{70.0} & \textbf{77.7} \\
\bottomrule

\end{tabular}
}
\end{table}
\else
\begin{table*}[!t]
\centering
\captionsetup{font={small,stretch=1},skip=3pt}
\linespread{0.95}\footnotesize
\renewcommand{\arraystretch}{0.94}
\setlength{\aboverulesep}{1.3pt}\setlength{\belowrulesep}{1.3pt}
\caption{UniVTAC success rates (SR, \%) and ranks across six tasks. Ties share a rank. Methods are ranked by mean SR in the last column group. TSB denotes Tactile Simple Baseline with a ResiP head. Throughout the tables, $\dagger$ denotes previously reported results and bold indicates the best scores within each setting.}
\label{tab:single_embodiment}
\setlength{\tabcolsep}{1.4pt}
\begin{tabularx}{\linewidth}{@{}l|*{6}{CC|}CC@{}}
\toprule\noalign{\vskip 1.2pt}
\multirow{2}{*}{Method} & \multicolumn{2}{c|}{\begin{tabular}[c]{@{}c@{}}Insert Hole\end{tabular}} & \multicolumn{2}{c|}{\begin{tabular}[c]{@{}c@{}}Insert Tube\end{tabular}} & \multicolumn{2}{c|}{\begin{tabular}[c]{@{}c@{}}Lift Bottle\end{tabular}} & \multicolumn{2}{c|}{\begin{tabular}[c]{@{}c@{}}Lift Can\end{tabular}} & \multicolumn{2}{c|}{\begin{tabular}[c]{@{}c@{}}Pull Out Key\end{tabular}} & \multicolumn{2}{c|}{\begin{tabular}[c]{@{}c@{}}Put Bottle\\in Shelf\end{tabular}} & \multicolumn{2}{c}{\begin{tabular}[c]{@{}c@{}}Avg.\end{tabular}} \\
\cmidrule{2-15}
 & SR$\uparrow$ & Rank$\downarrow$ & SR$\uparrow$ & Rank$\downarrow$ & SR$\uparrow$ & Rank$\downarrow$ & SR$\uparrow$ & Rank$\downarrow$ & SR$\uparrow$ & Rank$\downarrow$ & SR$\uparrow$ & Rank$\downarrow$ & SR$\uparrow$ & Rank$\downarrow$ \\
\midrule
$\pi_{0.5}$ & 27.0 & 8 & 37.0 & 9 & 91.0 & 6 & 80.0 & 3 & 42.0 & 8 & 36.0 & 8 & 52.2 & 6 \\
FTP-1$^\dagger$ & 64.0 & 5 & 79.0 & 2 & 97.0 & 2 & 65.0 & 6 & 48.0 & 6 & 47.0 & 5 & 66.7 & 4 \\
ManiFeel$^\dagger$ & 32.5 & 7 & 45.0 & 7 & 55.0 & 10 & 10.0 & 9 & \textbf{95.0} & \textbf{1} & 40.0 & 7 & 46.3 & 7 \\
RDP$^\dagger$ & 23.0 & 9 & 75.0 & 4 & 84.0 & 8 & 12.0 & 8 & 18.0 & 10 & 41.0 & 6 & 42.2 & 8 \\
ViTaR$^\dagger$ & --- & --- & 69.0 & 6 & 88.0 & 7 & 44.0 & 7 & 55.0 & 5 & 35.0 & 9 & --- & ---  \\
VT-MUSE$^\dagger$ & 68.0 & 4 & --- & --- & 84.0 & 8 & --- & --- & 38.0 & 9 & --- & --- & --- & ---  \\
\mbox{$\pi_{0.5}$ + LingBot-Touch Tiny} & 60.0 & 6 & 45.0 & 7 & 95.0 & 3 & 73.0 & 4 & 48.0 & 6 & \textbf{65.0} & \textbf{1} & 64.3 & 5 \\
SimpleTouch & 98.0 & 2 & 78.0 & 3 & 94.0 & 5 & \textbf{83.0} & \textbf{1} & 59.0 & 4 & 52.0 & 3 & 77.3 & 3 \\
\midrule
TSB & 98.0 & 2 & 75.0 & 4 & 95.0 & 3 & 68.0 & 5 & \textbf{95.0} & \textbf{1} & 50.0 & 4 & 80.2 & 2 \\
\rowcolor{retacgray}
\textbf{ReTac} & \textbf{100.0} & \textbf{1} & \textbf{90.0} & \textbf{1} & \textbf{100.0} & \textbf{1} & 82.0 & 2 & \textbf{95.0} & \textbf{1} & 60.0 & 2 & \textbf{87.8} & \textbf{1} \\
\bottomrule
\end{tabularx}
\par\vspace{5pt}
\caption{Cross-embodiment success rates (\%) without target-embodiment fine-tuning.}
\label{tab:cross_embodiment}
\footnotesize
\setlength{\tabcolsep}{3pt}
\begin{tabularx}{\linewidth}{@{}ll*{6}{C}c@{}}
\toprule\noalign{\vskip 1.2pt}
Transfer & Method & Insert Hole & Insert Tube & Lift Bottle & Lift Can & Pull Out Key & \begin{tabular}[c]{@{}c@{}}Put Bottle\\in Shelf\end{tabular} & Avg. \\
\midrule
\multirow{4}{*}{Franka $\to$ UR5e} & SimpleTouch & \textbf{98.0} & 80.0 & 23.0 & 15.0 & 20.0 & 43.0 & 46.5 \\
 & Fuse & 93.0 & 75.0 & 30.0 & 38.0 & 63.0 & 33.0 & 55.3 \\
 & TSB & \textbf{98.0} & 70.0 & 10.0 & 35.0 & 50.0 & 43.0 & 51.0 \\
 & \cellcolor{retacgray}ReTac & \cellcolor{retacgray}\textbf{98.0} & \cellcolor{retacgray}\textbf{83.0} & \cellcolor{retacgray}\textbf{48.0} & \cellcolor{retacgray}\textbf{44.0} & \cellcolor{retacgray}\textbf{73.0} & \cellcolor{retacgray}\textbf{50.0} & \cellcolor{retacgray}\textbf{66.0} \\
\midrule
\multirow{4}{*}{\shortstack[l]{Panda gripper\\$\to$ Allegro hand}} & SimpleTouch & \textbf{93.0} & \textbf{80.0} & 85.0 & 63.0 & 23.0 & 38.0 & 63.7 \\
 & Fuse & 80.0 & 78.0 & 93.0 & 75.0 & 23.0 & 63.0 & 68.7 \\
 & TSB & \textbf{93.0} & \textbf{80.0} & 70.0 & 75.0 & \textbf{53.0} & \textbf{70.0} & 73.5 \\
 & \cellcolor{retacgray}ReTac & \cellcolor{retacgray}85.0 & \cellcolor{retacgray}\textbf{80.0} & \cellcolor{retacgray}\textbf{98.0} & \cellcolor{retacgray}\textbf{83.0} & \cellcolor{retacgray}50.0 & \cellcolor{retacgray}\textbf{70.0} & \cellcolor{retacgray}\textbf{77.7} \\
\bottomrule
\end{tabularx}
\end{table*}
\fi

Modality weights enter the action query's attention as a bias:
\begin{equation}
O_{\mathrm{act}}=\mathrm{softmax}\!\left(
\frac{Q_{\mathrm{act}}K^\top}{\sqrt d}+M+b\right)V.
\label{eq:plan_attention}
\end{equation}
Here, $M$ is the main policy's attention mask. The bias is $\log w_V$ at visual tokens, $\log w_T$ at tactile tokens, and zero elsewhere. This bias controls the relative influence of vision and touch on action generation, allowing planning to rely more on reliable tactile feedback under visual degradation or cross-embodiment misalignment.

\textbf{Execution Action Validity Expert.} Every five steps, Exec AVE combines current feedback with the remaining plan through
\[
x_t=\left[v_{t_0},z_t,\Delta z_t,\bar z_t,\Delta q_t,D_t,p_t\right].
\]
The entries are cached visual features, current touch, tactile changes since planning, mean tactile history, proprioceptive changes, the next five targets relative to the current state, and normalized plan age, respectively. $u_t$ contains tactile prediction errors, validity flags, and prediction horizons for both fingertips. An MLP fuses the inputs and feeds an operation branch and a scaling-gain branch:
\begin{equation}
\begin{gathered}
h_t=\mathrm{MLP}\!\left(\mathrm{Concat}(x_t,u_t)\right),\\
m_t=\mathrm{Gate}(h_t)\in\{\mathrm K,\mathrm C,\mathrm R\},\quad
\alpha_t=\mathrm{Scale}(h_t).
\end{gathered}
\label{eq:alpha_gate}
\end{equation}
Exec AVE outputs the operation $m_t$ and an adaptive residual scaling gain $\alpha_t \in [0, 1.5]$. Specifically, $\mathrm{Gate}(h_t)$ predicts operation utilities $(\hat g_{\mathrm C}, \hat g_{\mathrm R})$ and applies a penalty-based selection rule, $s_{\mathrm K}=0$, $s_{\mathrm C}=\hat g_{\mathrm C}-\tau_{\mathrm C}$, and $s_{\mathrm R}=\hat g_{\mathrm R}-\tau_{\mathrm R}$, choosing $m_t=\arg\max_{m\in\{\mathrm K,\mathrm C,\mathrm R\}} s_m$. The thresholds $(\tau_{\mathrm C},\tau_{\mathrm R})$ are calibrated offline per task on validation demonstrations by minimizing five-step demonstration action RMSE subject to call-rate constraints that keep raw REPLAN proposals below $20\%$ and any single operation below $95\%$, without using closed-loop test success rates. In practice, Exec AVE actively regulates KEEP/CORRECT timing and adaptive residual gain $\alpha_t$ across all tasks, while REPLAN acts as a conservative high-cost fallback, as analyzed in Supplementary Secs.~C.2 and~E.1.

\subsection{Training Procedure}
\label{sec:training}
ReTac is trained in three stages using only offline demonstrations. First, we jointly fine-tune $\pi_{0.5}$ and Plan AVE with an action flow-matching loss~\cite{lipman2023flow} to learn planning and modality weights, retaining future tactile prediction as an auxiliary task following SimpleTouch:
\begin{equation}
\mathcal L_{\pi+\mathrm{Plan}}=
\mathcal L_{\mathrm{action}}^{\mathrm{FM}}+
\lambda_{\mathrm{touch}}\mathcal L_{\mathrm{touch}}^{\mathrm{FM}}.
\label{eq:plan_loss}
\end{equation}
Second, we freeze the planner, cache replanning disagreements, and train the correction head with Eq.~\eqref{eq:replan_target}. Finally, we freeze both the planner and correction head to train Exec AVE. The operation branch learns each candidate operation's normalized reduction in demonstration action error relative to keeping the plan, termed operation utility; the scaling-gain branch is supervised by the candidate coefficient $\alpha_t^\star \in \{0.5, 0.75, 1.0, 1.25, 1.5\}$ with the lowest correction error:
\begin{equation}
\mathcal L_{\mathrm{Exec}}=\mathcal L_{\mathrm{gain}}+
\mathrm{SmoothL1}_{\mathrm{valid}}\!\left(\alpha_t,\alpha_t^\star\right).
\label{eq:exec_loss}
\end{equation}
Here, $\mathcal L_{\mathrm{gain}}$ is the Smooth L1 regression loss for predicted operation utilities $(\hat g_{\mathrm C}, \hat g_{\mathrm R})$, distinct from the continuous residual scaling gain $\alpha_t$. Only Gate is supervised when the candidate residual norm is near zero. Correction directions are self-distilled from the policy's own updated plans, while visual and tactile modality weighting, operation selection, and correction magnitude are learned through demonstration supervision.

\section{Experiments}
Section~\ref{sec:experimental_setup} describes the data, implementation, and training setup. Section~\ref{sec:simulation_results} reports results on two simulation benchmarks, cross-embodiment transfer, and a comparison with a general-purpose agent. Section~\ref{sec:ablation} presents component and hyperparameter ablations, while Supplementary Secs.~C.1 and~C.2 isolate Plan AVE and Exec AVE separately. Section~\ref{sec:real_world} evaluates four physical manipulation tasks.

\iflatexml
\begin{table}[t]
\centering
\caption{ReTac and Astra on UniVTAC (10 trials per task). Latency (s) is per API request for Astra and per module call for ReTac.}
\label{tab:astra_comparison}
{\footnotesize
\textbf{(a) Successful trials and overall success rate}\par\smallskip
\begin{tabular}{lcccccccc}
\toprule
Method & \shortstack{Lift\\Bottle} & \shortstack{Pull Out\\Key} & \shortstack{Lift\\Can} & \shortstack{Put Bottle\\in Shelf} & \shortstack{Insert\\Hole} & \shortstack{Insert\\Tube} & Total & SR \\
\midrule
Astra (V+T) & \textbf{10} & \textbf{9} & \textbf{10} & \textbf{9} & \textbf{10} & 9 & \textbf{57} & \textbf{95.0\%} \\
Astra (V) & \textbf{10} & 3 & 8 & \textbf{9} & 6 & \textbf{10} & 46 & 76.7\% \\
\midrule
ReTac & \textbf{10} & 8 & 7 & 6 & \textbf{10} & \textbf{10} & 51 & 85.0\% \\
\bottomrule

\end{tabular}\par\medskip\textbf{(b) Latency (s)}\par\smallskip
\begin{tabular}{lccccc}
\toprule
Method & Operation & Mean & P50 & P90 & P99 \\
\midrule
Astra (V+T) & API request & $2.11 \times 10^{1}$ & $1.40 \times 10^{1}$ & $4.40 \times 10^{1}$ & $9.66 \times 10^{1}$ \\
Astra (V) & API request & $2.28 \times 10^{1}$ & $1.54 \times 10^{1}$ & $4.63 \times 10^{1}$ & $1.107 \times 10^{2}$ \\
\midrule
ReTac & Planning & $1.20 \times 10^{-1}$ & $1.18 \times 10^{-1}$ & $1.33 \times 10^{-1}$ & $1.64 \times 10^{-1}$ \\
ReTac & Tactile correction & $8.32 \times 10^{-3}$ & $7.70 \times 10^{-3}$ & $1.13 \times 10^{-2}$ & $1.52 \times 10^{-2}$ \\
\bottomrule

\end{tabular}
}
\end{table}

\begin{table}[t]
\centering
\caption{Comparison on the OpenViTac benchmark (SR, \%).}
\label{tab:openvitac-results}
{\footnotesize
\begin{tabular}{lcccc}
\toprule
Method & \shortstack[c]{Gear\\Assembly} & \shortstack[c]{Grasp\\Chip} & \shortstack[c]{Insert\\USB} & Avg. \\
\midrule
$\pi_{0.5}^{\dagger}$ & 87.0 & 39.0 & 14.0 & 46.7 \\
OpenVTLA$^{\dagger}$ & 91.0 & 69.0 & 27.0 & 62.3 \\
SimpleTouch & 85.0 & \textbf{77.5} & 17.5 & 60.0 \\
TSB & 85.0 & 57.5 & 15.0 & 52.5 \\
TSB + RD & 90.0 & 60.0 & 17.5 & 55.8 \\
\midrule
\textbf{ReTac} & \textbf{92.5} & 70.0 & \textbf{30.0} & \textbf{64.2} \\
\bottomrule

\end{tabular}
}
\end{table}

\begin{table}[t]
\centering
\caption{Ablation of residual injection gains on UniVTAC ($\Delta k=5$; SR, \%).}
\label{tab:gain_ablation}
{\footnotesize
\begin{tabular}{lcccccccc}
\toprule
Setting & $\alpha$ & \shortstack{Insert\\Hole} & \shortstack{Insert\\Tube} & \shortstack{Lift\\Bottle} & \shortstack{Lift\\Can} & \shortstack{Pull Out\\Key} & \shortstack{Put Bottle\\in Shelf} & Avg. \\
\midrule
\multirow{5}{*}{TSB + RD} & 0.5 & 95.0 & 83.0 & \textbf{100.0} & 73.0 & \textbf{95.0} & \textbf{65.0} & 85.2 \\
 & 0.75 & 95.0 & 78.0 & \textbf{100.0} & 78.0 & 88.0 & 48.0 & 81.2 \\
 & 1.0 & 99.0 & \textbf{90.0} & \textbf{100.0} & 80.0 & 85.0 & 56.0 & 85.0 \\
 & 1.25 & 98.0 & 80.0 & 95.0 & \textbf{83.0} & \textbf{95.0} & 58.0 & 84.8 \\
 & 1.5 & 90.0 & 83.0 & 90.0 & 78.0 & 90.0 & 48.0 & 79.8 \\
\midrule
\multicolumn{1}{@{}c}{ReTac} & Adaptive & \textbf{100.0} & \textbf{90.0} & \textbf{100.0} & 82.0 & \textbf{95.0} & 60.0 & \textbf{87.8} \\
\bottomrule

\end{tabular}
}
\end{table}
\else
\begin{table*}[t]
\centering
\captionsetup{font={small,stretch=1},skip=3pt}
\linespread{0.95}\footnotesize
\renewcommand{\arraystretch}{0.94}
\setlength{\aboverulesep}{1.3pt}\setlength{\belowrulesep}{1.3pt}
\caption{ReTac and Astra on UniVTAC (10 trials per task). Latency (s) is per API request for Astra and per module call for ReTac.}
\label{tab:astra_comparison}
\footnotesize
\setlength{\tabcolsep}{1pt}
\resizebox{\textwidth}{!}{%
\begin{tabular}{@{}l*{8}{c}l*{4}{c}@{}}
\toprule\noalign{\vskip 1.2pt}
\multirow{2}{*}{Method} & Lift & Pull Out & Lift & Put Bottle & Insert & Insert & \multicolumn{2}{c}{Overall} & \multirow{2}{*}{Operation} & \multirow{2}{*}{Mean} & \multirow{2}{*}{P50} & \multirow{2}{*}{P90} & \multirow{2}{*}{P99} \\
\cmidrule(lr){8-9}
 & Bottle & Key & Can & in Shelf & Hole & Tube & Total & SR & & & & & \\
\midrule
Astra (V+T) & \textbf{10} & \textbf{9} & \textbf{10} & \textbf{9} & \textbf{10} & 9 & \textbf{57} & \textbf{95.0\%} & API request & $2.11\!\times\!10^{1}$ & $1.40\!\times\!10^{1}$ & $4.40\!\times\!10^{1}$ & $9.66\!\times\!10^{1}$ \\
Astra (V) & \textbf{10} & 3 & 8 & \textbf{9} & 6 & \textbf{10} & 46 & 76.7\% & API request & $2.28\!\times\!10^{1}$ & $1.54\!\times\!10^{1}$ & $4.63\!\times\!10^{1}$ & $1.107\!\times\!10^{2}$ \\
\midrule
\multirow{2}{*}{ReTac} & \multirow{2}{*}{\textbf{10}} & \multirow{2}{*}{8} & \multirow{2}{*}{7} & \multirow{2}{*}{6} & \multirow{2}{*}{\textbf{10}} & \multirow{2}{*}{\textbf{10}} & \multirow{2}{*}{51} & \multirow{2}{*}{85.0\%} & Planning & $1.20\!\times\!10^{-1}$ & $1.18\!\times\!10^{-1}$ & $1.33\!\times\!10^{-1}$ & $1.64\!\times\!10^{-1}$ \\
 & & & & & & & & & Tactile correction & $8.32\!\times\!10^{-3}$ & $7.70\!\times\!10^{-3}$ & $1.13\!\times\!10^{-2}$ & $1.52\!\times\!10^{-2}$ \\
\bottomrule
\end{tabular}}
\par\vspace{5pt}
\noindent\begin{minipage}[t]{\columnwidth}\vspace{0pt}
\centering
\captionsetup{type=table,font={small,stretch=1},skip=3pt}
\linespread{0.95}\footnotesize
\renewcommand{\arraystretch}{1.05}
\setlength{\aboverulesep}{1.3pt}\setlength{\belowrulesep}{1.3pt}
\caption{Comparison on the OpenViTac benchmark (SR, \%).}
\label{tab:openvitac-results}
\setlength{\tabcolsep}{3pt}
\renewcommand{\tabularxcolumn}[1]{m{#1}}
\begin{tabularx}{\linewidth}{@{}>{\raggedright\arraybackslash}m{0.25\linewidth}CCCC@{}}
\toprule\noalign{\vskip 1.2pt}
{\centering Method\par} & \shortstack[c]{Gear\\Assembly} & \shortstack[c]{Grasp\\Chip} & \shortstack[c]{Insert\\USB} & Avg. \\
\midrule
$\pi_{0.5}^{\dagger}$ & 87.0 & 39.0 & 14.0 & 46.7 \\
OpenVTLA$^{\dagger}$ & 91.0 & 69.0 & 27.0 & 62.3 \\
SimpleTouch & 85.0 & \textbf{77.5} & 17.5 & 60.0 \\
TSB & 85.0 & 57.5 & 15.0 & 52.5 \\
TSB + RD & 90.0 & 60.0 & 17.5 & 55.8 \\
\midrule
\rowcolor{retacgray}\textbf{ReTac} & \textbf{92.5} & 70.0 & \textbf{30.0} & \textbf{64.2} \\
\bottomrule
\end{tabularx}
\end{minipage}\hfill
\begin{minipage}[t]{\columnwidth}\vspace{0pt}
\centering
\captionsetup{font={small,stretch=1},skip=3pt}
\linespread{0.95}\footnotesize
\renewcommand{\arraystretch}{0.94}
\setlength{\aboverulesep}{1.3pt}\setlength{\belowrulesep}{1.3pt}
\caption{Ablation of residual injection gains on UniVTAC ($\Delta k=5$; SR, \%).}
\label{tab:gain_ablation}
\footnotesize
\setlength{\tabcolsep}{1.7pt}
\newcommand{\gainhead}[2]{\begin{tabular}[c]{@{}c@{}}#1\\#2\end{tabular}}
\resizebox{\linewidth}{!}{%
\begin{tabular}{@{}c|c|cccccc|c@{}}
\toprule\noalign{\vskip 1.2pt}
Setting & $\alpha$ & \gainhead{Insert}{Hole} & \gainhead{Insert}{Tube} & \gainhead{Lift}{Bottle} & \gainhead{Lift}{Can} & \gainhead{Pull Out}{Key} & \gainhead{Put Bottle}{in Shelf} & Avg. \\
\midrule
\multirow{5}{*}{TSB + RD} & 0.5 & 95.0 & 83.0 & \textbf{100.0} & 73.0 & \textbf{95.0} & \textbf{65.0} & 85.2 \\
 & 0.75 & 95.0 & 78.0 & \textbf{100.0} & 78.0 & 88.0 & 48.0 & 81.2 \\
 & 1.0 & 99.0 & \textbf{90.0} & \textbf{100.0} & 80.0 & 85.0 & 56.0 & 85.0 \\
 & 1.25 & 98.0 & 80.0 & 95.0 & \textbf{83.0} & \textbf{95.0} & 58.0 & 84.8 \\
 & 1.5 & 90.0 & 83.0 & 90.0 & 78.0 & 90.0 & 48.0 & 79.8 \\
\midrule
\multicolumn{1}{@{}c}{ReTac} & Adaptive & \textbf{100.0} & \textbf{90.0} & \textbf{100.0} & 82.0 & \textbf{95.0} & 60.0 & \textbf{87.8} \\
\bottomrule
\end{tabular}%
}
\end{minipage}
\end{table*}
\fi

\subsection{Experimental Setup}
\label{sec:experimental_setup}
\noindent\textbf{Data and evaluation.}
Following SimpleTouch~\cite{yang2026simpletouch} and UniVTAC~\cite{chen2026univtac}, the main experiments train task-specific models per task, while Supplementary Sec.~D.1 evaluates a cross-task shared RD head. On each of the six UniVTAC tasks, 50 demonstrations are split into 40 training trajectories and 10 held-out offline evaluation trajectories; for Exec AVE, trajectories 0--31 train the network and trajectories 32--39 calibrate $(\tau_{\mathrm C},\tau_{\mathrm R})$ offline, disjoint from all closed-loop test seeds. Standard evaluation runs 100 trials per task on UniVTAC and 40 trials per task on OpenViTac~\cite{wu2026openvitac} with 100 demonstrations per task, both operating at $60\,\mathrm{Hz}$. We report per-task success rates and their unweighted mean. Zero-shot transfer evaluates Franka to UR5e and Panda gripper to Allegro hand via a thumb--index pinch lookup mapping, retaining all source weights and calibrated thresholds without target fine-tuning. The Astra comparison uses the original seeds over 10 trials per task~\cite{yang2026astra}.

\noindent\textbf{Implementation and training.}
A frozen tactile encoder samples features every five steps with four cached history frames. The main policy runs 10 Euler steps to predict 50-step chunks and executes at most 25 steps. Fuse uses tactile history alone; TSB adds a ResiP head with a five-step interval and unit gain; TSB + RD replaces ResiP with RD; TSB + AVE adds both AVE branches to TSB; and ReTac combines RD with both AVE branches, evaluating every five steps with offline validation-calibrated thresholds $(\tau_{\mathrm C},\tau_{\mathrm R})$ and adaptive residual scaling gains bounded in $[0, 1.5]$. For Franka-to-UR5e transfer, forward and inverse kinematics map joint targets through end-effector poses.

All variants share the same $\pi_{0.5}$ initialization and are trained on 16 A800 or PRO6000 GPUs with a global batch size of 128 for 10,000 updates, using 500 warmup steps to $5\times10^{-5}$ and cosine decay to $5\times10^{-6}$. With the planner frozen, simulation ResiP heads train via online PPO under a 40-episode budget, whereas RD trains offline for 30 epochs and Exec AVE trains for 100 epochs with AdamW at learning rate $10^{-3}$, weight decay $10^{-4}$, and batch size 128.

\iflatexml
\begin{figure*}[t]
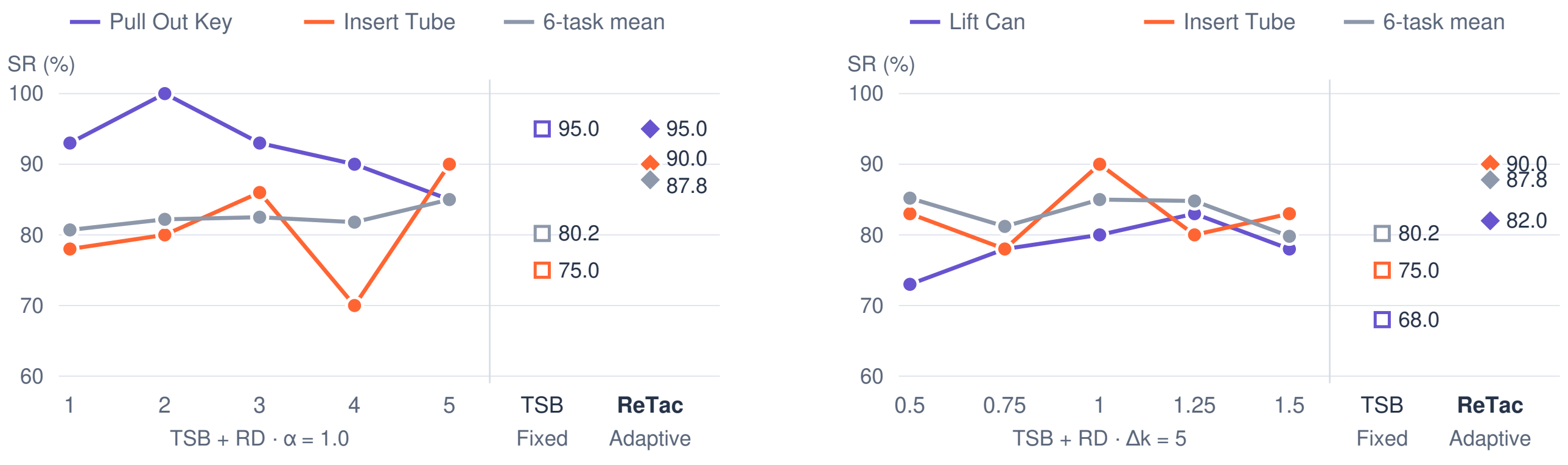

\centering
\begin{tabular}{cc}
\includegraphics[width=0.48\textwidth]{correction-interval-html.svg} &
\includegraphics[width=0.48\textwidth]{injection-gain-html.svg}\\
(a) Correction interval ($\alpha=1$). & (b) Residual injection gain ($\Delta k=5$).
\end{tabular}
\caption{Sensitivity to correction timing and strength. TSB + RD uses fixed intervals and gains; TSB and ReTac are shown at their respective configurations.}
\label{fig:correction-sensitivity}
\end{figure*}
\else
\begin{figure*}[!t]
\centering
\begin{minipage}[t]{\columnwidth}
\centering
\iflatexml
\includegraphics[width=\linewidth]{correction-interval-html.svg}
\else
\includegraphics[width=\linewidth]{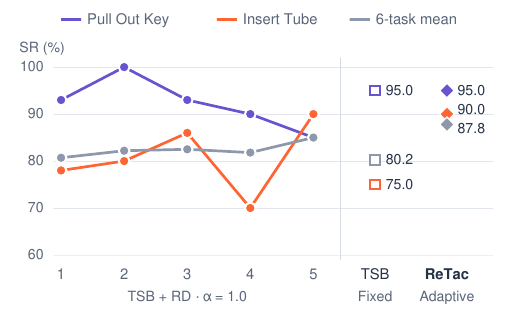}
\fi
\caption*{(a) Correction interval ($\alpha=1$).}
\end{minipage}\hfill
\begin{minipage}[t]{\columnwidth}
\centering
\iflatexml
\includegraphics[width=\linewidth]{injection-gain-html.svg}
\else
\includegraphics[width=\linewidth]{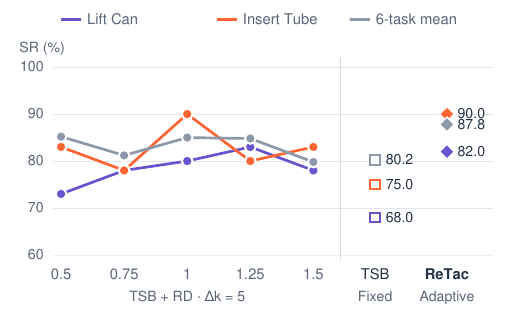}
\fi
\caption*{(b) Residual injection gain ($\Delta k=5$).}
\end{minipage}
\caption{Sensitivity to correction timing and strength. TSB + RD uses fixed intervals and gains; TSB and ReTac are shown at their respective configurations.}
\label{fig:correction-sensitivity}
\end{figure*}
\fi

\subsection{Simulation Results}
\label{sec:simulation_results}
\noindent\textbf{UniVTAC.}
As shown in Table~\ref{tab:single_embodiment}, ReTac achieves the highest mean success rate of $87.8\%$, compared with $77.3\%$ for SimpleTouch and $80.2\%$ for TSB. It leads or ties on four of six tasks. Table~\ref{tab:component_ablation} further shows that replacing the ResiP head with RD raises TSB's mean success from $80.2\%$ to $85.0\%$; combining RD with AVE further improves it to $87.8\%$, also exceeding TSB + AVE at $81.5\%$. These comparisons support the complementary roles of learning corrective actions from updated plans and adapting their use to the current interaction.

Relative to TSB + RD in Table~\ref{tab:gain_ablation}, ReTac raises success from $85.0\%$ to $95.0\%$ on Pull Out Key and from $56.0\%$ to $60.0\%$ on Put Bottle in Shelf. This broader coverage supports coordinated observation weighting and selective intervention: the planner emphasizes reliable sensory evidence, while the execution expert regulates departures from the reference trajectory. Together, these mechanisms help reconcile local contact correction with global trajectory stability across different manipulation demands.

\noindent\textbf{OpenViTac.}
Table~\ref{tab:openvitac-results} shows that ReTac ranks first with $64.2\%$ mean success, achieving the best results on Gear Assembly at $92.5\%$ and Insert USB at $30.0\%$. Under the 40-episode sparse-reward budget, adding an uncalibrated fixed-gain PPO head in TSB degrades performance relative to SimpleTouch from $60.0\%$ to $52.5\%$, particularly on the fragile Grasp Chip task where success drops from $77.5\%$ to $57.5\%$, illustrating the risk of disruptive residual perturbations. With the 8-D continuous gripper residual extension applied identically to ResiP and RD on Grasp Chip, ReTac improves all three tasks over TSB + RD, raising Insert USB from $17.5\%$ to $30.0\%$ and Grasp Chip from $60.0\%$ to $70.0\%$. These gains support contact-conditioned intervention across tasks with different tolerances for corrective motion, improving both precise alignment and delicate handling over fixed residual policies.

\noindent\textbf{Zero-shot transfer.}
Embodiment changes alter action-to-motion mappings and contact geometry. A correction effective on the source robot may therefore become insufficient or disruptive on the target; indeed, on Franka $\to$ UR5e, adding the 40-episode PPO residual head in TSB lowers mean success relative to history-only Fuse from $55.3\%$ to $51.0\%$. Table~\ref{tab:cross_embodiment} shows that ReTac achieves the highest mean success in both transfer settings: $66.0\%$ from Franka to UR5e and $77.7\%$ from the Panda gripper to the Allegro hand under the thumb--index pinch mapping, while SimpleTouch achieves $46.5\%$ and $63.7\%$ and TSB reaches $51.0\%$ and $73.5\%$, respectively. ReTac also outperforms the history-only Fuse variant on every task in both settings, indicating that its transfer advantage extends beyond retaining temporal contact information.

On UR5e, key extraction rises from $20.0\%$ with SimpleTouch to $73.0\%$; with Allegro, bottle placement improves from $38.0\%$ to $70.0\%$. We attribute this adaptability to coupling planning-time reliability assessment with execution-time correction: Plan AVE adjusts the influence of sensory evidence under embodiment shift, while Exec AVE uses current feedback to regulate whether and how strongly to revise the trajectory. These gains support contact-aware action regulation for zero-shot transfer across arms and grippers.

\noindent\textbf{Multimodal agent.}
Across 60 trials with matched scene seeds, ReTac achieves $85.0\%$ success, compared with $76.7\%$ for vision-only Astra in Table~\ref{tab:astra_comparison}. Astra with vision and touch achieves $95.0\%$, with a mean latency of $21.13\,\mathrm{s}$ per API request. ReTac takes $120.35\,\mathrm{ms}$ per local planning call and $8.32\,\mathrm{ms}$ per tactile encoding plus correction call, or $12.73\,\mathrm{ms}$ when including AVE decision-making. The fast correction path lets the robot respond to contact between planning calls at low inference cost.

\iflatexml
\begin{table}[t]
\centering
\caption{Component ablations on UniVTAC. RD denotes the replanning-disagreement-distilled correction head.}
\label{tab:component_ablation}
{\footnotesize
\begin{tabular}{lccc}
\toprule
Name & RD & AVE & Avg. (\%) \\
\midrule
Tactile Simple Baseline &  &  & 80.2 \\
\midrule
\#1 & $\checkmark$ &  & 85.0 \\
\#2 &  & $\checkmark$ & 81.5 \\
\textbf{\#3} & $\checkmark$ & $\checkmark$ & \textbf{87.8} \\
\bottomrule

\end{tabular}
}
\end{table}
\else
\begin{table}[!t]
\centering
\captionsetup{font={small,stretch=1},skip=3pt}
\linespread{0.95}\footnotesize
\renewcommand{\arraystretch}{0.94}
\setlength{\aboverulesep}{1.3pt}\setlength{\belowrulesep}{1.3pt}
\caption{Component ablations on UniVTAC. RD denotes the replanning-disagreement-distilled correction head.}
\label{tab:component_ablation}
\footnotesize
\setlength{\tabcolsep}{2.5pt}
\renewcommand{\tabularxcolumn}[1]{m{#1}}
\begin{tabularx}{\columnwidth}{@{}>{\centering\arraybackslash}m{0.42\columnwidth}|CC|C@{}}
\toprule\noalign{\vskip 1.2pt}
Name & RD & AVE & Avg. (\%) \\
\midrule
Tactile Simple Baseline & & & 80.2 \\
\midrule
\#1 & $\checkmark$ & & 85.0 \\
\#2 & & $\checkmark$ & 81.5 \\
\textbf{\#3} & $\checkmark$ & $\checkmark$ & \textbf{87.8} \\
\bottomrule
\end{tabularx}
\end{table}
\fi

\iflatexml
\begin{table}[t]
\centering
\caption{Ablation of correction intervals on UniVTAC ($\alpha=1.0$; SR, \%).}
\label{tab:interval_ablation}
{\footnotesize
\begin{tabular}{lcccccccc}
\toprule
Setting & $\Delta k$ & \shortstack{Insert\\Hole} & \shortstack{Insert\\Tube} & \shortstack{Lift\\Bottle} & \shortstack{Lift\\Can} & \shortstack{Pull Out\\Key} & \shortstack{Put Bottle\\in Shelf} & Avg. \\
\midrule
\multirow{5}{*}{TSB + RD} & 1 & 90.0 & 78.0 & \textbf{100.0} & 78.0 & 93.0 & 45.0 & 80.7 \\
 & 2 & 90.0 & 80.0 & 95.0 & 78.0 & \textbf{100.0} & 50.0 & 82.2 \\
 & 3 & 93.0 & 86.0 & 90.0 & 78.0 & 93.0 & 55.0 & 82.5 \\
 & 4 & 98.0 & 70.0 & \textbf{100.0} & \textbf{83.0} & 90.0 & 50.0 & 81.8 \\
 & 5 & 99.0 & \textbf{90.0} & \textbf{100.0} & 80.0 & 85.0 & 56.0 & 85.0 \\
\midrule
\multicolumn{1}{@{}c}{ReTac} & 5 & \textbf{100.0} & \textbf{90.0} & \textbf{100.0} & 82.0 & 95.0 & \textbf{60.0} & \textbf{87.8} \\
\bottomrule

\end{tabular}
}
\end{table}
\else
\begin{table}[!t]
\centering
\captionsetup{font={small,stretch=1},skip=3pt}
\linespread{0.95}\footnotesize
\renewcommand{\arraystretch}{0.94}
\setlength{\aboverulesep}{1.3pt}\setlength{\belowrulesep}{1.3pt}
\caption{Ablation of correction intervals on UniVTAC ($\alpha=1.0$; SR, \%).}
\label{tab:interval_ablation}
\setlength{\tabcolsep}{1.7pt}
\newcommand{\intervalhead}[2]{\begin{tabular}[c]{@{}c@{}}#1\\#2\end{tabular}}
\resizebox{\linewidth}{!}{%
\begin{tabular}{@{}c|c|cccccc|c@{}}
\toprule\noalign{\vskip 1.2pt}
Setting & $\Delta k$ & \intervalhead{Insert}{Hole} & \intervalhead{Insert}{Tube} & \intervalhead{Lift}{Bottle} & \intervalhead{Lift}{Can} & \intervalhead{Pull Out}{Key} & \intervalhead{Put Bottle}{in Shelf} & Avg. \\
\midrule
\multirow{5}{*}{TSB + RD} & 1 & 90.0 & 78.0 & \textbf{100.0} & 78.0 & 93.0 & 45.0 & 80.7 \\
 & 2 & 90.0 & 80.0 & 95.0 & 78.0 & \textbf{100.0} & 50.0 & 82.2 \\
 & 3 & 93.0 & 86.0 & 90.0 & 78.0 & 93.0 & 55.0 & 82.5 \\
 & 4 & 98.0 & 70.0 & \textbf{100.0} & \textbf{83.0} & 90.0 & 50.0 & 81.8 \\
 & 5 & 99.0 & \textbf{90.0} & \textbf{100.0} & 80.0 & 85.0 & 56.0 & 85.0 \\
\midrule
\multicolumn{1}{@{}c}{ReTac} & 5 & \textbf{100.0} & \textbf{90.0} & \textbf{100.0} & 82.0 & 95.0 & \textbf{60.0} & \textbf{87.8} \\
\bottomrule
\end{tabular}%
}
\end{table}
\fi

\subsection{Ablation Study}
\label{sec:ablation}
We examine the need for adaptive intervention, the RD correction head, and AVE, followed by an alternative fusion design.

\noindent\textbf{Correction intervals and injection gains.}
We vary TSB + RD's correction interval at $\alpha=1$ and injection gain at $\Delta k=5$ in Tables~\ref{tab:interval_ablation} and~\ref{tab:gain_ablation}. Figure~\ref{fig:correction-sensitivity} shows task-dependent optima: key extraction favors a two-step interval, whereas tube insertion favors five; can lifting and tube insertion favor gains of $1.25$ and $1.0$, respectively. More frequent or stronger corrections therefore do not consistently improve performance. ReTac achieves $87.8\%$ mean success, exceeding every fixed configuration. These results support feedback-dependent correction timing and magnitude to balance responses to resistance with precise, stable motion.

\noindent\textbf{Replanning-disagreement-distilled head (RD).}
Replacing TSB's ResiP head with RD under the same planner and other settings increases mean success from $80.2\%$ to $85.0\%$ in Table~\ref{tab:component_ablation}. Under matched 40-seed evaluation, RD also outperforms Policy Decorator, ResiP, and CR-DAgger in mean success, as reported in Supplementary Sec.~A.1. Thus, the policy's revisions after receiving new observations provide effective correction supervision. The lightweight RD head learns these adjustments from offline demonstrations, reducing reliance on full replanning during execution.

\noindent\textbf{Action Validity Expert (AVE).}
Adding AVE to TSB and TSB + RD improves mean success from $80.2\%$ to $81.5\%$ and from $85.0\%$ to $87.8\%$, respectively, in Table~\ref{tab:component_ablation}, showing complementary benefits from correction and intervention regulation. With the planner and RD head frozen, the full two-level AVE also outperforms single-level execution-gating baselines including ReMoE, Gated Attention, and sLSTM with fixed planning weights in a system-level comparison in Supplementary Sec.~A.2, while controlled single-factor isolations of Plan AVE and Exec AVE are provided in Supplementary Secs.~C.1 and~C.2. Adding AVE to TSB + RD improves key extraction from $85.0\%$ to $95.0\%$ and bottle placement from $56.0\%$ to $60.0\%$. Key extraction must overcome resistance after alignment; bottle placement connects visual alignment with contact adjustment. Plan AVE weights vision and touch by observation validity, while Exec AVE selects whether and how strongly to correct. This regulation helps correct obstructed motion while preserving an aligned trajectory.

\begin{figure}[!t]
\centering
\iflatexml
\includegraphics[width=\linewidth]{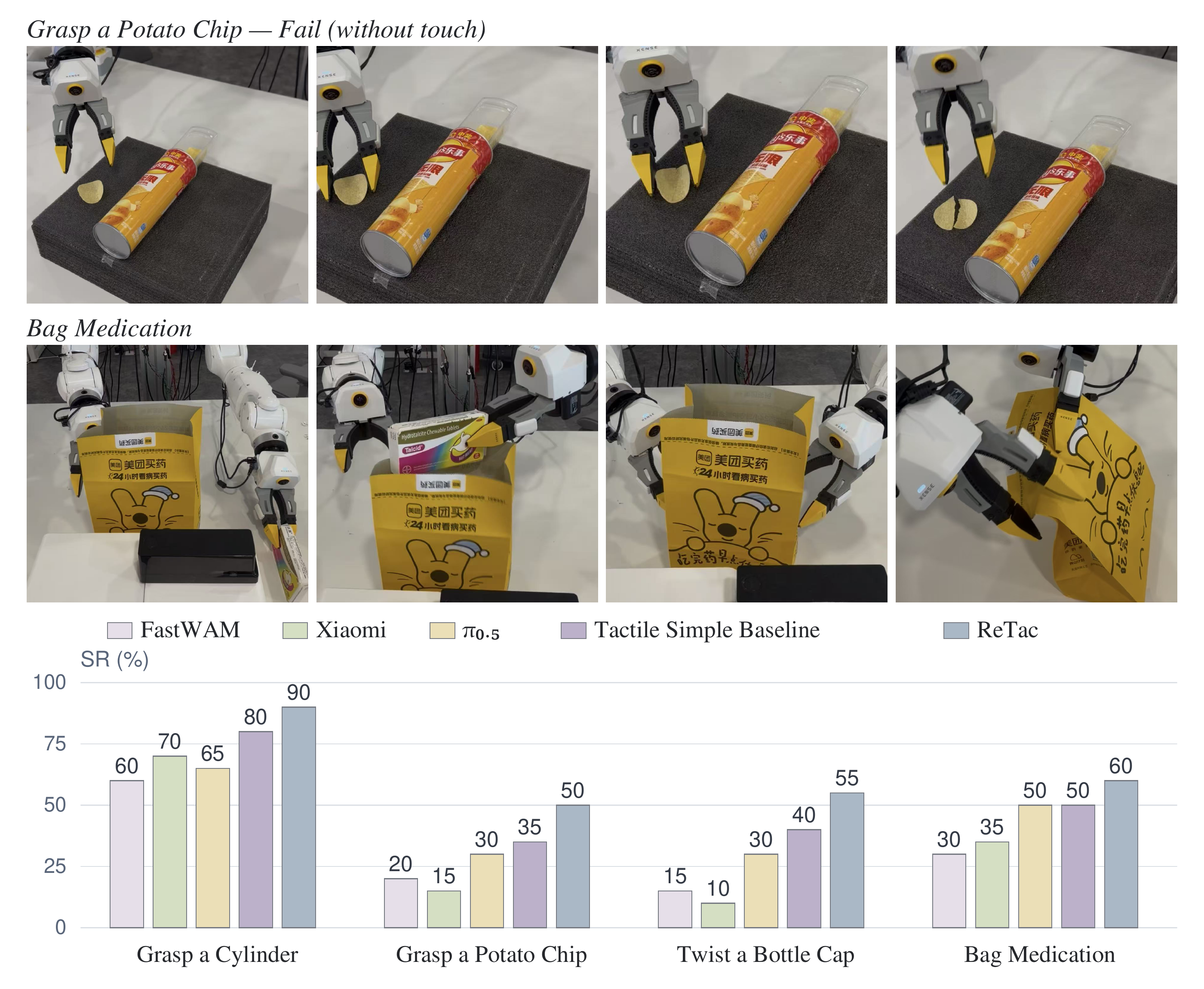}
\else
\includegraphics[width=\linewidth]{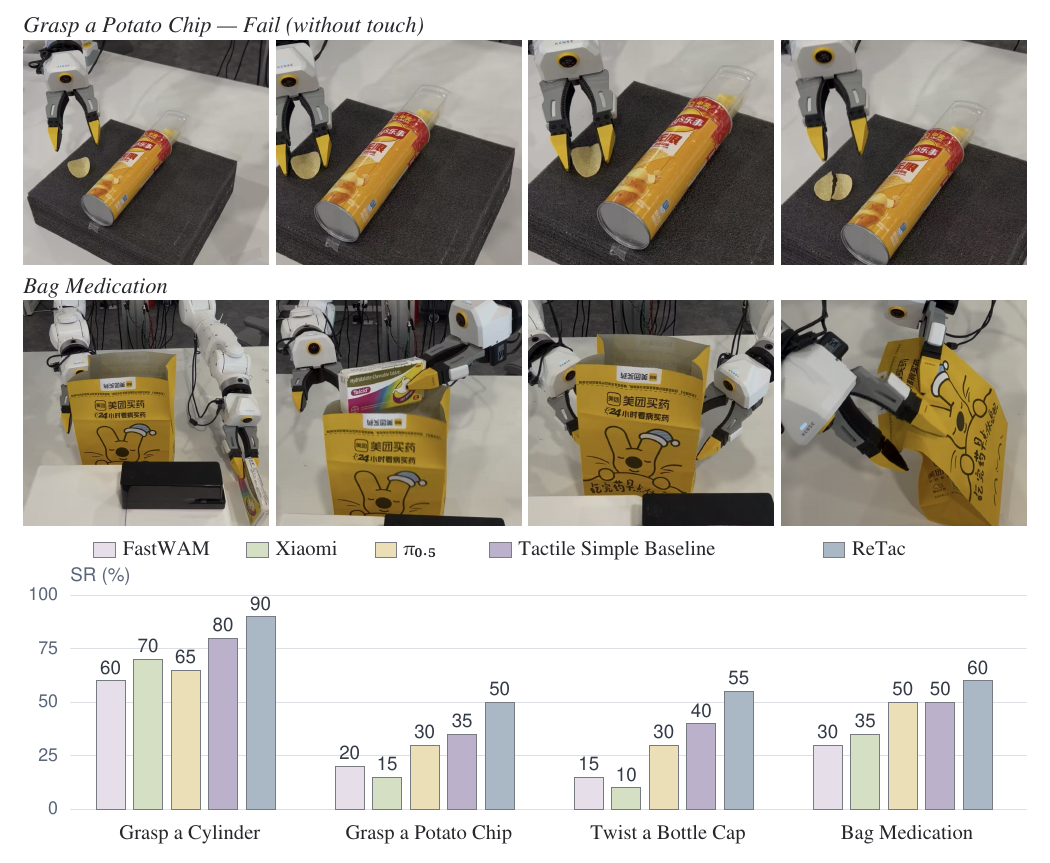}
\fi
\caption{Real-world contact-rich manipulation tasks and results.}
\label{fig:real-world}
\end{figure}

\noindent\textbf{Visual and tactile fusion.}
We compare SimpleTouch with full-attention fusion in Isaac Sim 4.5, letting the current tactile query attend jointly to tactile history and visual features. Mean success rises from $73.2\%$ to $76.2\%$ in Supplementary Sec.~A.3: can lifting improves from $55.0\%$ to $69.0\%$, but hole insertion and key extraction decline slightly. Broader feature interaction thus yields uneven benefits across manipulation demands, further motivating Plan AVE's explicit assessment of visual and tactile validity for state-dependent modality weighting.

\subsection{Real-World Evaluation}
\label{sec:real_world}

\noindent\textbf{Experimental setup.}
We evaluate four tasks on a UMI platform~\cite{chi2024umi} mounted on a dual-arm Tianji Marvin M6-CCS robot with Xense XTac UMI G1 grippers and Xense GS-PS fingertip visuotactile sensors synchronized at $50\,\mathrm{Hz}$: grasping and placing a cylinder into a basket, returning a potato chip intact into its box, twisting off and setting down a bottle cap, and bagging and sealing medication. Sensor sampling rates, single- versus dual-arm configurations, and demonstration counts are detailed in Supplementary Sec.~E.3. Over 20 trials per task, we compare visuotactile Fast-WAM~\cite{yuan2026fastwam}, visuotactile Xiaomi XR-1~\cite{xiaomi2026xr1}, vision-only $\pi_{0.5}$, real-world TSB, and ReTac fine-tuned on the same demonstrations. To avoid unsafe physical PPO, real-world TSB retains the ResiP head with fixed unit-gain execution under offline replanning-disagreement supervision, and both TSB and ReTac adopt the 8-D continuous gripper residual on potato-chip grasping.

\noindent\textbf{Results analysis.}
As shown in Figure~\ref{fig:real-world}, ReTac achieves the highest success rate on all four tasks, averaging $63.8\%$, compared with $51.3\%$ for real-world TSB, $43.8\%$ for $\pi_{0.5}$, $32.5\%$ for Xiaomi XR-1, and $31.3\%$ for Fast-WAM. Success increases over TSB from $80.0\%$ to $90.0\%$ for cylinder grasping, from $35.0\%$ to $50.0\%$ for potato-chip grasping, from $40.0\%$ to $55.0\%$ for cap twisting, and from $50.0\%$ to $60.0\%$ for medication bagging. These gains span distinct contact constraints: responding to slip, avoiding fracture, maintaining contact during rotation, and coordinating alignment, placement, and sealing. The potato-chip sequence without touch in Figure~\ref{fig:real-world} shows a breakage failure, illustrating that executing a grasp does not ensure successful fragile-object handling. Plan AVE coordinates visual localization with tactile feedback, the 8-D residual head enables continuous aperture adjustment upon contact, and Exec AVE regulates intervention timing and magnitude, suppressing unnecessary perturbations once a stable grasp is established. Consistent gains across tasks demonstrate ReTac's ability to accommodate stable grasping, fragile-object manipulation, and sequential contact adjustment, translating contact feedback into effective closed-loop control.

\section{Conclusion}

In this work, we introduce \textbf{MXI-ReTac}, a reliability-aware visuotactile framework for contact-rich manipulation. Specifically, MXI-ReTac distills overlapping replanning disagreements into a lightweight tactile residual head. This head enables fast within-chunk correction. During planning, its Action Validity Expert assesses observation reliability to weight visual and tactile attention. During execution, it estimates correction utility to regulate intervention timing and residual strength. Training relies solely on standard offline demonstrations. It requires neither online reinforcement learning nor manual recovery data. We evaluate MXI-ReTac on simulated benchmarks, zero-shot cross-embodiment transfer, and real-world UMI tasks. It consistently outperforms vision-only and fixed-schedule tactile baselines with millisecond-scale local inference.

\clearpage
\iflatexml
{\normalsize
\bibliographystyle{ieeenat_fullname}
\bibliography{main}
}
\else
{\small
\bibliographystyle{ieeenat_fullname}
\bibliography{main}
}
\fi

\clearpage

\raggedbottom
\setlength{\textfloatsep}{3.3pt}
\setlength{\dbltextfloatsep}{5.5pt}
\setlength{\floatsep}{2.7pt}
\setlength{\dblfloatsep}{2.7pt}
\setlength{\intextsep}{3.3pt}
\setcounter{dbltopnumber}{1}
\renewcommand{\dbltopfraction}{0.85}
\afterpage{\afterpage{\setcounter{dbltopnumber}{3}}}
\twocolumn[\begin{center}\Large\bfseries Supplementary Material\end{center}\vspace{6pt}]

\appendix
\renewcommand\thesection{\Alph{section}}
\numberwithin{table}{section}
\numberwithin{equation}{section}

This supplement extends the main paper following the section order cited in Sec.~4.3: Section~\ref{sec:supp_ablations} compares correction heads, gating systems, and feature fusion; Section~\ref{sec:supp_supervision} contrasts demonstration and replanning supervision; Sections~\ref{sec:supp_plan_ave} and~\ref{sec:supp_exec_ave} isolate Plan AVE and Exec AVE; Section~\ref{sec:supp_shared} evaluates a cross-task shared RD head; and Section~\ref{sec:supp_details} details formulations, threshold calibration, transfer interfaces, real-world setup, and latency.

\section{Additional Ablations}
\label{sec:supp_ablations}

\subsection{Correction-Head Comparison}
\label{sec:supp_head_comp}

With the Fuse planner and execution settings fixed, we compare Policy Decorator~\cite{yuan2025policy}, ResiP~\cite{ankile2024resip}, CR-DAgger~\cite{xu2025crdagger}, and RD under a matched budget of 40 training episodes and 10 validation episodes per task. All four heads are evaluated on the exact same 40 disjoint test seeds per task in Table~\ref{tab:supp_head_comp}(a); 100-trial results for ResiP and RD appear in Tables~1 and~5 of the main paper. Parameters remain frozen during validation and testing without contributing training data, while physical interaction counts vary across methods because episodes terminate early on success or failure. This 40-episode cap matches the 40 offline demonstrations used by RD, evaluating low-budget sample efficiency rather than the asymptotic ceiling of extended online RL.

\iflatexml
\begin{table}[t]
\centering
\caption*{Per-task collection limits for 40 training episodes; actual counts depend on early termination.}
{\footnotesize
\begin{tabular}{lcc}
\toprule
\shortstack{Task} & \shortstack{Physical\\actions} & \shortstack{Five-step\\samples} \\
\midrule
\shortstack[l]{Insert Hole, Insert Tube,\\Lift Can, Pull Out Key (each)} & 12,000 & 2,400 \\
Lift Bottle & 20,000 & 4,000 \\
Put Bottle in Shelf & 32,000 & 6,400 \\
\bottomrule

\end{tabular}
}
\end{table}
\else
\begin{table}[H]
\centering
\small
\setlength{\tabcolsep}{3pt}
\renewcommand{\arraystretch}{1.08}
\renewcommand{\tabularxcolumn}[1]{m{#1}}
\caption*{Per-task collection limits for 40 training episodes; actual counts depend on early termination.}
\begin{tabularx}{\columnwidth}{@{}>{\raggedright\arraybackslash}Xcc@{}}
\toprule
\supphead{Task} & \supphead{Physical\\actions} & \supphead{Five-step\\samples} \\
\midrule
Insert Hole, Insert Tube, Lift Can, Pull Out Key (each) & 12,000 & 2,400 \\
\addlinespace[2pt]
Lift Bottle & 20,000 & 4,000 \\
Put Bottle in Shelf & 32,000 & 6,400 \\
\bottomrule
\end{tabularx}
\end{table}
\fi

\paragraph{Baseline setups.}
\textbf{Policy Decorator} trains an online SAC residual policy over the frozen planner, recording one transition every five steps to output a $5 \times 7$ residual sequence under task success rewards, with 32-transition warm-up, batch size 128, learning rate $10^{-4}$, and two updates per eight transitions. \textbf{CR-DAgger} runs four 10-episode rounds, querying the frozen planner every five steps to form time-aligned five-step residual labels and retraining for 30 epochs on cumulative data after each round. \textbf{ResiP} trains online with PPO as in the main paper, while \textbf{RD} distills offline time-aligned planner revisions without environment interaction.

\iflatexml
\begin{table}[t]
\centering
\caption{Correction-head and gating comparisons on UniVTAC (SR, \%). (a) Correction heads. (b) Gating configurations.}
\label{tab:supp_head_comp}
{\footnotesize
\begin{tabular}{lcccccccc}
\toprule
\shortstack{Method} & \shortstack{$n$} & \shortstack{Insert\\Hole} & \shortstack{Insert\\Tube} & \shortstack{Lift\\Bottle} & \shortstack{Lift\\Can} & \shortstack{Pull Out\\Key} & \shortstack{Put Bottle\\in Shelf} & \shortstack{Avg.} \\
\midrule
\multicolumn{9}{@{}l}{\textit{(a) Correction heads}} \\
Policy Decorator & 40 & 92.5 & 80.0 & 97.5 & 72.5 & 90.0 & 27.5 & 76.7 \\
ResiP & 40 & 97.5 & 75.0 & 95.0 & 67.5 & \textbf{95.0} & 50.0 & 80.0 \\
CR-DAgger & 40 & 95.0 & 72.5 & 97.5 & 72.5 & 87.5 & 40.0 & 77.5 \\
RD & 40 & \textbf{100.0} & \textbf{87.5} & \textbf{100.0} & \textbf{77.5} & \textbf{95.0} & \textbf{55.0} & \textbf{85.8} \\
\midrule
\multicolumn{9}{@{}l}{\textit{(b) Gating configurations}} \\
ReMoE & 40 & 95.0 & 82.5 & 97.5 & 72.5 & 85.0 & 47.5 & 80.0 \\
Gated Attention & 40 & 90.0 & \textbf{90.0} & 92.5 & 67.5 & 77.5 & 37.5 & 75.8 \\
sLSTM & 40 & 82.5 & 72.5 & 97.5 & 77.5 & 90.0 & 35.0 & 75.8 \\
AVE & 40 & \textbf{100.0} & 87.5 & \textbf{100.0} & \textbf{82.5} & \textbf{95.0} & \textbf{60.0} & \textbf{87.5} \\
\bottomrule

\end{tabular}
}
\end{table}
\else
\begin{table*}[t]
\captionsetup{skip=3pt}
\centering
\small
\setlength{\tabcolsep}{5pt}
\renewcommand{\arraystretch}{1.08}
\caption{Correction-head and gating comparisons on UniVTAC (SR, \%). (a) Correction heads. (b) Gating configurations.}
\label{tab:supp_head_comp}
\label{tab:supp_gating}
\begin{tabular*}{\textwidth}{@{\extracolsep{\fill}}lcccccccc@{}}
\toprule
\supphead{Method} & \supphead{$n$} & \supphead{Insert\\Hole} & \supphead{Insert\\Tube} & \supphead{Lift\\Bottle} & \supphead{Lift\\Can} & \supphead{Pull Out\\Key} & \supphead{Put Bottle\\in Shelf} & \supphead{Avg.} \\
\midrule
\multicolumn{9}{@{}l}{\textit{(a) Correction heads}} \\
Policy Decorator & 40 & 92.5 & 80.0 & 97.5 & 72.5 & 90.0 & 27.5 & 76.7 \\
ResiP & 40 & 97.5 & 75.0 & 95.0 & 67.5 & \textbf{95.0} & 50.0 & 80.0 \\
CR-DAgger & 40 & 95.0 & 72.5 & 97.5 & 72.5 & 87.5 & 40.0 & 77.5 \\
RD & 40 & \textbf{100.0} & \textbf{87.5} & \textbf{100.0} & \textbf{77.5} & \textbf{95.0} & \textbf{55.0} & \textbf{85.8} \\
\midrule
\multicolumn{9}{@{}l}{\textit{(b) Gating configurations}} \\
ReMoE & 40 & 95.0 & 82.5 & 97.5 & 72.5 & 85.0 & 47.5 & 80.0 \\
Gated Attention & 40 & 90.0 & \textbf{90.0} & 92.5 & 67.5 & 77.5 & 37.5 & 75.8 \\
sLSTM & 40 & 82.5 & 72.5 & 97.5 & 77.5 & 90.0 & 35.0 & 75.8 \\
AVE & 40 & \textbf{100.0} & 87.5 & \textbf{100.0} & \textbf{82.5} & \textbf{95.0} & \textbf{60.0} & \textbf{87.5} \\
\bottomrule
\end{tabular*}
\end{table*}
\fi

RD reaches the highest mean success rate of 85.8\%, or 206 of 240 trials, ahead of ResiP at 80.0\%, CR-DAgger at 77.5\%, and Policy Decorator at 76.7\%, ranking first or tied on all six tasks. Interval and gain sweeps appear in Fig.~3 and Tables~5 and~7 of the main paper.

\subsection{Gating Comparison}
\label{sec:supp_gating}

With the planner and RD head frozen, we compare full two-level AVE against single-level execution gates based on ReMoE~\cite{wang2025remoe}, Gated Attention~\cite{qiu2025gated}, and sLSTM~\cite{beck2024xlstm} with fixed planning weights. The execution inputs comprise cached visual features, current touch, tactile change, mean tactile history, joint change, the next five action offsets, and plan age; the first four groups are projected separately before fusion with the remaining state features and $u_t$. Each alternative gate $G$ predicts
\begin{equation}
\begin{gathered}
h_t = G(\mathrm{Concat}(x_t, u_t)), \\
m_t = \arg\max_{m \in \{K, C, R\}} [\mathrm{Linear}_3(h_t)]_m, \\
\alpha_t = 2\,\sigma(\mathrm{Linear}_1(h_t)),
\label{eq:alt_gate}
\end{gathered}
\end{equation}
with sLSTM resetting its recurrent state each episode. Alternative gates output scaling gains in $[0, 2]$, whereas ReTac bounds $\alpha_t \in [0, 1.5]$. Both Exec AVE and all alternative gates use the exact same offline trajectory split per task: trajectories 0--31 for training, 32--39 for threshold calibration or checkpoint selection, and 40--49 for held-out offline evaluation, disjoint from the 40 matched closed-loop test seeds. Alternative gates train for 100 epochs using AdamW, learning rate $10^{-3}$, weight decay $10^{-4}$, and a batch of four episodes.

AVE averages 87.5\% versus 80.0\% for ReMoE and 75.8\% for Gated Attention and sLSTM, leading on five tasks and by 12.5 points on Put Bottle in Shelf. Because Table~\ref{tab:supp_gating}(b) compares full two-level AVE with bounded gain $\alpha_t \in [0, 1.5]$ against single-level gates, Tables~\ref{tab:supp_fixed_weights} and~\ref{tab:supp_timing_gain} isolate Plan AVE and Exec AVE individually.

\subsection{Full-Attention Fusion}
\label{sec:supp_full_attn}

In Isaac Sim 4.5, we test letting the current tactile query attend jointly to tactile history and visual features, reproducing SimpleTouch~\cite{yang2026simpletouch} under the same setup. For vision-only $\pi_{0.5}$, Insert Hole and Insert Tube are reproduced and the remaining four entries come from Table~1 of SimpleTouch~\cite{yang2026simpletouch}. Full attention raises the mean from 73.2\% to 76.2\% in Table~\ref{tab:supp_full_attn}, driven mainly by Lift Can from 55.0\% to 69.0\%, while Insert Hole and Pull Out Key dip slightly.

\iflatexml
\begin{table}[t]
\centering
\caption{Alternative feature fusion in Isaac Sim 4.5 (SR, \%). $\dagger$ denotes values from SimpleTouch~\cite{yang2026simpletouch}, Table~1.}
\label{tab:supp_full_attn}
{\footnotesize
\begin{tabular}{lccccccc}
\toprule
\shortstack{Method} & \shortstack{Insert\\Hole} & \shortstack{Insert\\Tube} & \shortstack{Lift\\Bottle} & \shortstack{Lift\\Can} & \shortstack{Pull Out\\Key} & \shortstack{Put Bottle\\in Shelf} & \shortstack{Avg.} \\
\midrule
$\pi_{0.5}$ & 25.0 & 77.0 & $92.0^{\dagger}$ & $59.0^{\dagger}$ & $36.0^{\dagger}$ & $31.0^{\dagger}$ & 53.3 \\
SimpleTouch & \textbf{83.0} & 98.0 & \textbf{100.0} & 55.0 & \textbf{64.0} & 39.0 & 73.2 \\
Full attention & 82.0 & \textbf{100.0} & \textbf{100.0} & \textbf{69.0} & 63.0 & \textbf{43.0} & \textbf{76.2} \\
\bottomrule

\end{tabular}
}
\end{table}
\else
\begin{table*}[t]
\vspace*{4pt} 
\captionsetup{skip=3pt}
\centering
\small
\setlength{\tabcolsep}{5pt}
\caption{Alternative feature fusion in Isaac Sim 4.5 (SR, \%). $\dagger$ denotes values from SimpleTouch~\cite{yang2026simpletouch}, Table~1.}
\label{tab:supp_full_attn}
\begin{tabular*}{\textwidth}{@{\extracolsep{\fill}}lccccccc@{}}
\toprule
\supphead{Method} & \supphead{Insert\\Hole} & \supphead{Insert\\Tube} & \supphead{Lift\\Bottle} & \supphead{Lift\\Can} & \supphead{Pull Out\\Key} & \supphead{Put Bottle\\in Shelf} & \supphead{Avg.} \\
\midrule
$\pi_{0.5}$ & 25.0 & 77.0 & $92.0^{\dagger}$ & $59.0^{\dagger}$ & $36.0^{\dagger}$ & $31.0^{\dagger}$ & 53.3 \\
SimpleTouch & \textbf{83.0} & 98.0 & \textbf{100.0} & 55.0 & \textbf{64.0} & 39.0 & 73.2 \\
Full attention & 82.0 & \textbf{100.0} & \textbf{100.0} & \textbf{69.0} & 63.0 & \textbf{43.0} & \textbf{76.2} \\
\bottomrule
\end{tabular*}
\end{table*}
\fi

\section{Further Analysis of Correction Supervision}
\label{sec:supp_supervision}

\subsection{Demonstration versus Replanning Supervision}
\label{sec:supp_demo_vs_rd}

Following Sec.~3.3 of the main paper, for a reference chunk $A$ at $t_0$ with $H = 50$ and $N = 25$, each five-step correction time $t \in \mathcal{T}_{t_0} = \{t_0+5, t_0+10, \dots, t_0+N\}$ has offset $g = t - t_0$ and replanned chunk $B^{(t)}$. With $W = 5$, we compare RD against direct demonstration-residual supervision over the same time-aligned joint targets at steps $t + j$:
\begin{equation}
\begin{gathered}
\Delta a_{t}^{\star} = \frac{1}{5}\sum_{j=0}^{4}\big(u_{t,j}^{\mathrm{arm}} - A_{g+j}^{\mathrm{arm}}\big), \\
u_{t,j} = \begin{cases}
a_{t+j}^{\mathrm{demo}}, & \text{demonstration supervision,} \\
B_{j}^{(t)}, & \text{RD supervision.}
\end{cases}
\label{eq:demo_vs_rd}
\end{gathered}
\end{equation}
Both Smooth L1 inputs are normalized by per-dimension training scales. Because both $A$ and $B^{(t)}$ are evaluated on offline demonstration observations rather than closed-loop rollouts, RD distills the planner's observation-driven revision direction rather than closed-loop recovery error, while Exec AVE regulates execution-time intervention. Keeping the planner and Plan AVE fixed, we freeze each trained correction head and retrain Exec AVE for it before evaluating on the same 40 test seeds per task.

\iflatexml
\begin{table}[t]
\centering
\caption{Correction supervision on UniVTAC (SR, \%).}
\label{tab:supp_supervision}
{\footnotesize
\begin{tabular}{lcccccccc}
\toprule
\shortstack{Target} & \shortstack{$n$} & \shortstack{Insert\\Hole} & \shortstack{Insert\\Tube} & \shortstack{Lift\\Bottle} & \shortstack{Lift\\Can} & \shortstack{Pull Out\\Key} & \shortstack{Put Bottle\\in Shelf} & \shortstack{Avg.} \\
\midrule
Demonstration residual & 40 & 95.0 & \textbf{92.5} & 95.0 & 72.5 & 90.0 & \textbf{60.0} & 84.2 \\
\rowcolor{retacgray}
RD (ReTac) & 40 & \textbf{100.0} & 87.5 & \textbf{100.0} & \textbf{82.5} & \textbf{95.0} & \textbf{60.0} & \textbf{87.5} \\
\bottomrule

\end{tabular}
}
\end{table}
\else
\begin{table*}[t]
\captionsetup{skip=3pt}
\centering
\small
\setlength{\tabcolsep}{4.5pt}
\caption{Correction supervision on UniVTAC (SR, \%).}
\label{tab:supp_supervision}
\begin{tabular*}{\textwidth}{@{\extracolsep{\fill}}lcccccccc@{}}
\toprule
\supphead{Target} & \supphead{$n$} & \supphead{Insert\\Hole} & \supphead{Insert\\Tube} & \supphead{Lift\\Bottle} & \supphead{Lift\\Can} & \supphead{Pull Out\\Key} & \supphead{Put Bottle\\in Shelf} & \supphead{Avg.} \\
\midrule
Demonstration residual & 40 & 95.0 & \textbf{92.5} & 95.0 & 72.5 & 90.0 & \textbf{60.0} & 84.2 \\
RD (ReTac) & 40 & \textbf{100.0} & 87.5 & \textbf{100.0} & \textbf{82.5} & \textbf{95.0} & \textbf{60.0} & \textbf{87.5} \\
\bottomrule
\end{tabular*}
\end{table*}
\fi

RD improves mean success from 84.2\% to 87.5\%, or 202 to 210 of 240 trials, outperforming demonstration residuals on four tasks including a 10-point margin on Lift Can, while trailing only on Insert Tube.

\section{Adaptive Modality Weighting and Intervention}
\label{sec:supp_ave}

\subsection{Adaptive versus Fixed Modality Weights}
\label{sec:supp_plan_ave}

To isolate Plan AVE, we replace its outputs at inference with fixed modality weights $(w_V, w_T)$ entering Eq.~(4) of the main paper as $\log w_V$ and $\log w_T$, keeping the planner, RD head, and Exec AVE unchanged across the same 40 test seeds per task.

\iflatexml
\begin{table}[t]
\centering
\caption{Fixed planning weights $(w_V, w_T)$ versus adaptive ReTac on UniVTAC (SR, \%).}
\label{tab:supp_fixed_weights}
{\footnotesize
\begin{tabular}{lcccccccc}
\toprule
\shortstack{$(w_V, w_T)$} & \shortstack{$n$} & \shortstack{Insert\\Hole} & \shortstack{Insert\\Tube} & \shortstack{Lift\\Bottle} & \shortstack{Lift\\Can} & \shortstack{Pull Out\\Key} & \shortstack{Put Bottle\\in Shelf} & \shortstack{Avg.} \\
\midrule
$(0.75, 1)$ & 40 & 95.0 & 90.0 & 97.5 & 80.0 & 92.5 & \textbf{62.5} & 86.3 \\
$(1, 0.75)$ & 40 & 95.0 & 77.5 & \textbf{100.0} & 77.5 & 90.0 & 52.5 & 82.1 \\
$(1.5, 0.5)$ & 40 & 82.5 & 60.0 & 95.0 & 80.0 & 92.5 & 47.5 & 76.3 \\
$(0.5, 1.5)$ & 40 & 95.0 & 82.5 & 97.5 & 65.0 & 92.5 & 50.0 & 80.4 \\
$(2, 1)$ & 40 & 97.5 & 80.0 & 92.5 & 77.5 & 87.5 & 52.5 & 81.3 \\
$(1, 2)$ & 40 & 92.5 & \textbf{97.5} & \textbf{100.0} & 72.5 & 85.0 & 47.5 & 82.5 \\
$(1, 1)$ & 40 & 97.5 & 90.0 & \textbf{100.0} & \textbf{85.0} & 87.5 & 55.0 & 85.8 \\
\rowcolor{retacgray}
ReTac (adaptive) & 40 & \textbf{100.0} & 87.5 & \textbf{100.0} & 82.5 & \textbf{95.0} & 60.0 & \textbf{87.5} \\
\bottomrule

\end{tabular}
}
\end{table}
\else
\begin{table*}[t]
\vspace*{4pt} 
\captionsetup{skip=3pt}
\centering
\small
\setlength{\tabcolsep}{4.5pt}
\caption{Fixed planning weights $(w_V, w_T)$ versus adaptive ReTac on UniVTAC (SR, \%).}
\label{tab:supp_fixed_weights}
\begin{tabular*}{\textwidth}{@{\extracolsep{\fill}}lcccccccc@{}}
\toprule
\supphead{$(w_V, w_T)$} & \supphead{$n$} & \supphead{Insert\\Hole} & \supphead{Insert\\Tube} & \supphead{Lift\\Bottle} & \supphead{Lift\\Can} & \supphead{Pull Out\\Key} & \supphead{Put Bottle\\in Shelf} & \supphead{Avg.} \\
\midrule
$(0.75, 1)$ & 40 & 95.0 & 90.0 & 97.5 & 80.0 & 92.5 & \textbf{62.5} & 86.3 \\
$(1, 0.75)$ & 40 & 95.0 & 77.5 & \textbf{100.0} & 77.5 & 90.0 & 52.5 & 82.1 \\
$(1.5, 0.5)$ & 40 & 82.5 & 60.0 & 95.0 & 80.0 & 92.5 & 47.5 & 76.3 \\
$(0.5, 1.5)$ & 40 & 95.0 & 82.5 & 97.5 & 65.0 & 92.5 & 50.0 & 80.4 \\
$(2, 1)$ & 40 & 97.5 & 80.0 & 92.5 & 77.5 & 87.5 & 52.5 & 81.3 \\
$(1, 2)$ & 40 & 92.5 & \textbf{97.5} & \textbf{100.0} & 72.5 & 85.0 & 47.5 & 82.5 \\
$(1, 1)$ & 40 & 97.5 & 90.0 & \textbf{100.0} & \textbf{85.0} & 87.5 & 55.0 & 85.8 \\
ReTac (adaptive) & 40 & \textbf{100.0} & 87.5 & \textbf{100.0} & 82.5 & \textbf{95.0} & 60.0 & \textbf{87.5} \\
\bottomrule
\end{tabular*}
\end{table*}
\fi

No fixed pair dominates across tasks in Table~\ref{tab:supp_fixed_weights}: $(1, 2)$ reaches 97.5\% on Insert Tube but drops to 72.5\% on Lift Can and 47.5\% on Put Bottle in Shelf, whereas equal weighting $(1, 1)$ recovers the latter two to 85.0\% and 55.0\% while lowering Insert Tube to 90.0\%. Adaptive ReTac achieves the highest mean of 87.5\%, ahead of the best fixed pairs $(0.75, 1)$ at 86.3\% and $(1, 1)$ at 85.8\%.

\subsection{Intervention Behavior}
\label{sec:supp_exec_ave}

\paragraph{Timing versus residual scaling.}
Using the shared RD head from Sec.~\ref{sec:supp_shared_head} with identical planner, Plan AVE, and 40 matched test seeds per task, we compare three execution modes: Configuration C uses full Exec AVE timing and adaptive $\alpha_t$, defining per-task call-rate targets for A and B; Configuration A applies fixed timing with $\alpha = 1$; Configuration B uses Exec AVE timing with $\alpha = 1$.

\iflatexml
\iflatexml
\begin{table}[t]
\centering
\caption{\strut Intervention timing and gain on UniVTAC (SR, \%; $n=40$; A/B/C: fixed/AVE/AVE timing, $\alpha=1/1/\alpha_t$).}
\label{tab:supp_timing_gain}
{\footnotesize
\begin{tabular}{lccccccc}
\toprule
\shortstack{Config.} & \shortstack{Insert\\Hole} & \shortstack{Insert\\Tube} & \shortstack{Lift\\Bottle} & \shortstack{Lift\\Can} & \shortstack{Pull Out\\Key} & \shortstack{Put Bottle\\in Shelf} & \shortstack{Avg.} \\
\midrule
\shortstack{A} & 95.0 & 87.5 & 95.0 & 77.5 & 92.5 & \textbf{57.5} & 84.2 \\
\shortstack{B} & 95.0 & \textbf{95.0} & \textbf{100.0} & 80.0 & 92.5 & 55.0 & 86.3 \\
\shortstack{C} & \textbf{97.5} & 92.5 & \textbf{100.0} & \textbf{82.5} & \textbf{97.5} & 55.0 & \textbf{87.5} \\
\bottomrule

\end{tabular}
}
\end{table}
\else
\begin{table}[t]
\caption{\strut Intervention timing and gain on UniVTAC (SR, \%; $n=40$; A/B/C: fixed/AVE/AVE timing, $\alpha=1/1/\alpha_t$).}
\label{tab:supp_timing_gain}
\begin{tabular}{@{}c|cccccc|c@{}}
\toprule\noalign{\vskip 1.2pt}
\supphead{Config.} & \supphead{Insert\\Hole} & \supphead{Insert\\Tube} & \supphead{Lift\\Bottle} & \supphead{Lift\\Can} & \supphead{Pull Out\\Key} & \supphead{Put Bottle\\in Shelf} & \supphead{Avg.} \\
\midrule
\supphead{A} & 95.0 & 87.5 & 95.0 & 77.5 & 92.5 & \textbf{57.5} & 84.2 \\
\supphead{B} & 95.0 & \textbf{95.0} & \textbf{100.0} & 80.0 & 92.5 & 55.0 & 86.3 \\
\supphead{C} & \textbf{97.5} & 92.5 & \textbf{100.0} & \textbf{82.5} & \textbf{97.5} & 55.0 & \textbf{87.5} \\
\bottomrule
\end{tabular}
\end{table}
\fi
\iflatexml
\begin{table}[t]
\centering
\caption{\strut UniVTAC call rates per 100 executed control steps for C; per-task rate targets for A and B.}
\label{tab:supp_call_rates}
{\footnotesize
\begin{tabular}{lccccccc}
\toprule
\shortstack{Metric} & \shortstack{Insert\\Hole} & \shortstack{Insert\\Tube} & \shortstack{Lift\\Bottle} & \shortstack{Lift\\Can} & \shortstack{Pull Out\\Key} & \shortstack{Put Bottle\\in Shelf} & \shortstack{All\\trajectories} \\
\midrule
\shortstack{Correct} & 15.00 & 11.37 & 14.39 & 5.08 & 10.47 & 15.10 & 12.78 \\
\shortstack{Replan} & 1.56 & 0.00 & 2.12 & 0.00 & 0.00 & 0.00 & 0.64 \\
\shortstack{Main-planner} & 5.17 & 4.30 & 5.34 & 4.18 & 4.27 & 4.03 & 4.51 \\
\bottomrule

\end{tabular}
}
\end{table}
\else
\begin{table}[t]
\caption{\strut UniVTAC call rates per 100 executed control steps for C; per-task rate targets for A and B.}
\label{tab:supp_call_rates}
\begin{tabular}{@{}c|cccccc|c@{}}
\toprule\noalign{\vskip 1.2pt}
\supphead{Metric} & \supphead{Insert\\Hole} & \supphead{Insert\\Tube} & \supphead{Lift\\Bottle} & \supphead{Lift\\Can} & \supphead{Pull Out\\Key} & \supphead{Put Bottle\\in Shelf} & \supphead{All\\trajectories} \\
\midrule
\supphead{Correct} & 15.00 & 11.37 & 14.39 & 5.08 & 10.47 & 15.10 & 12.78 \\
\supphead{Replan} & 1.56 & 0.00 & 2.12 & 0.00 & 0.00 & 0.00 & 0.64 \\
\supphead{Main-planner} & 5.17 & 4.30 & 5.34 & 4.18 & 4.27 & 4.03 & 4.51 \\
\bottomrule
\end{tabular}
\end{table}
\fi
\else
\iflatexml
\begin{table}[t]
\centering
\caption{\strut Intervention timing and gain on UniVTAC (SR, \%; $n=40$; A/B/C: fixed/AVE/AVE timing, $\alpha=1/1/\alpha_t$).}
\label{tab:supp_timing_gain}
{\footnotesize
\begin{tabular}{lccccccc}
\toprule
\shortstack{Config.} & \shortstack{Insert\\Hole} & \shortstack{Insert\\Tube} & \shortstack{Lift\\Bottle} & \shortstack{Lift\\Can} & \shortstack{Pull Out\\Key} & \shortstack{Put Bottle\\in Shelf} & \shortstack{Avg.} \\
\midrule
\shortstack{A} & 95.0 & 87.5 & 95.0 & 77.5 & 92.5 & \textbf{57.5} & 84.2 \\
\shortstack{B} & 95.0 & \textbf{95.0} & \textbf{100.0} & 80.0 & 92.5 & 55.0 & 86.3 \\
\shortstack{C} & \textbf{97.5} & 92.5 & \textbf{100.0} & \textbf{82.5} & \textbf{97.5} & 55.0 & \textbf{87.5} \\
\bottomrule

\end{tabular}
}
\end{table}

\begin{table}[t]
\centering
\caption{\strut UniVTAC call rates per 100 executed control steps for C; per-task rate targets for A and B.}
\label{tab:supp_call_rates}
{\footnotesize
\begin{tabular}{lccccccc}
\toprule
\shortstack{Metric} & \shortstack{Insert\\Hole} & \shortstack{Insert\\Tube} & \shortstack{Lift\\Bottle} & \shortstack{Lift\\Can} & \shortstack{Pull Out\\Key} & \shortstack{Put Bottle\\in Shelf} & \shortstack{All\\trajectories} \\
\midrule
\shortstack{Correct} & 15.00 & 11.37 & 14.39 & 5.08 & 10.47 & 15.10 & 12.78 \\
\shortstack{Replan} & 1.56 & 0.00 & 2.12 & 0.00 & 0.00 & 0.00 & 0.64 \\
\shortstack{Main-planner} & 5.17 & 4.30 & 5.34 & 4.18 & 4.27 & 4.03 & 4.51 \\
\bottomrule

\end{tabular}
}
\end{table}
\else
\begin{table*}[t]
\vspace*{4pt} 

\captionsetup{skip=3pt,justification=raggedright,singlelinecheck=false,strut=on,font={small,stretch=1},labelsep=period}
\noindent
\begin{minipage}[t]{\dimexpr(\textwidth-\columnsep)/2\relax}
\vspace{0pt}
\caption{\strut Intervention timing and gain on UniVTAC (SR, \%; $n=40$; A/B/C: fixed/AVE/AVE timing, $\alpha=1/1/\alpha_t$).}
\label{tab:supp_timing_gain}
\end{minipage}\hfill
\begin{minipage}[t]{\dimexpr(\textwidth-\columnsep)/2\relax}
\vspace{0pt}
\caption{\strut UniVTAC call rates per 100 executed control steps for C; per-task rate targets for A and B.}
\label{tab:supp_call_rates}
\end{minipage}
\par\nointerlineskip
\noindent
\begin{minipage}[t]{\dimexpr(\textwidth-\columnsep)/2\relax}
\vspace{0pt}
\centering
\linespread{0.95}\footnotesize
\renewcommand{\arraystretch}{1.12}
\setlength{\aboverulesep}{1.3pt}\setlength{\belowrulesep}{1.3pt}
\setlength{\tabcolsep}{1.7pt}
\resizebox{\linewidth}{!}{%
\begin{tabular}{@{}c|cccccc|c@{}}
\toprule\noalign{\vskip 1.2pt}
\supphead{Config.} & \supphead{Insert\\Hole} & \supphead{Insert\\Tube} & \supphead{Lift\\Bottle} & \supphead{Lift\\Can} & \supphead{Pull Out\\Key} & \supphead{Put Bottle\\in Shelf} & \supphead{Avg.} \\
\midrule
\supphead{A} & 95.0 & 87.5 & 95.0 & 77.5 & 92.5 & \textbf{57.5} & 84.2 \\
\supphead{B} & 95.0 & \textbf{95.0} & \textbf{100.0} & 80.0 & 92.5 & 55.0 & 86.3 \\
\supphead{C} & \textbf{97.5} & 92.5 & \textbf{100.0} & \textbf{82.5} & \textbf{97.5} & 55.0 & \textbf{87.5} \\
\bottomrule
\end{tabular}%
}
\end{minipage}\hfill
\begin{minipage}[t]{\dimexpr(\textwidth-\columnsep)/2\relax}
\vspace{0pt}
\centering
\linespread{0.95}\footnotesize
\renewcommand{\arraystretch}{1.12}
\setlength{\aboverulesep}{1.3pt}\setlength{\belowrulesep}{1.3pt}
\setlength{\tabcolsep}{1.7pt}
\resizebox{\linewidth}{!}{%
\begin{tabular}{@{}c|cccccc|c@{}}
\toprule\noalign{\vskip 1.2pt}
\supphead{Metric} & \supphead{Insert\\Hole} & \supphead{Insert\\Tube} & \supphead{Lift\\Bottle} & \supphead{Lift\\Can} & \supphead{Pull Out\\Key} & \supphead{Put Bottle\\in Shelf} & \supphead{All\\trajectories} \\
\midrule
\supphead{Correct} & 15.00 & 11.37 & 14.39 & 5.08 & 10.47 & 15.10 & 12.78 \\
\supphead{Replan} & 1.56 & 0.00 & 2.12 & 0.00 & 0.00 & 0.00 & 0.64 \\
\supphead{Main-planner} & 5.17 & 4.30 & 5.34 & 4.18 & 4.27 & 4.03 & 4.51 \\
\bottomrule
\end{tabular}%
}
\end{minipage}
\end{table*}
\fi
\fi

Successes rise monotonically from 202 to 207 and 210 out of 240 trials across A, B, and C in Table~\ref{tab:supp_timing_gain}, showing distinct gains from state-dependent timing at 84.2\% to 86.3\% and adaptive scaling to 87.5\%.

\paragraph{Call-rate profile and early-replanning sparsity.}
Table~\ref{tab:supp_call_rates} lists Configuration C's observed calls per 100 executed control steps, computed as 100 times total calls over total executed steps. Total main-planner calls include early replanning calls; the latter are a subset, not an additional category. Exec AVE actively regulates CORRECT timing at 5.08 to 15.10 calls per 100 steps and scaling $\alpha_t$ across all six tasks, whereas early REPLAN is deliberately sparse at 0.64 calls per 100 steps pooled. Under the calibrated penalties in Table~\ref{tab:supp_thresholds} and the sub-20\% raw proposal constraint, REPLAN acts as a conservative high-cost fallback, triggering on Insert Hole and Lift Bottle while staying inactive where five-step local residuals already minimize validation action error. Because offline supervision in Eq.~\eqref{eq:op_benefit} uses in-distribution demonstrations, severe OOD jams are underrepresented and make REPLAN conservative; moreover, per-step margins at $n=40$ have limited statistical power and motivate larger-scale paired testing.

\section{Sharing Corrections across Tasks}
\label{sec:supp_shared}

\subsection{Shared Correction Head and AVE Reuse}
\label{sec:supp_shared_head}

To test cross-task sharing, we train one joint RD head across all six UniVTAC tasks while keeping task-specific planners. The combined dataset contains 240 training and 60 validation trajectories, or 66,185 training and 15,980 validation windows. Uniformly sampling task, trajectory, and window prevents longer episodes from dominating; training runs for 30 epochs with AdamW, batch size 128, learning rate $10^{-3}$, and weight decay $10^{-4}$, selecting epoch 22 by mean task-wise validation RMSE.

\iflatexml
\iflatexml
\begin{table}[t]
\centering
\caption{\strut Shared RD head with and without reusing the original task-specific AVE on UniVTAC (SR, \%; $n=40$ trials per task).}
\label{tab:supp_shared}
{\footnotesize
\begin{tabular}{lccccccc}
\toprule
\shortstack{Setting} & \shortstack{Insert\\Hole} & \shortstack{Insert\\Tube} & \shortstack{Lift\\Bottle} & \shortstack{Lift\\Can} & \shortstack{Pull Out\\Key} & \shortstack{Put Bottle\\in Shelf} & \shortstack{Avg.} \\
\midrule
\shortstack{AVE off} & 95.0 & 87.5 & 95.0 & 80.0 & 92.5 & 52.5 & 83.8 \\
\shortstack{Original AVE} & \textbf{97.5} & \textbf{92.5} & \textbf{100.0} & \textbf{82.5} & \textbf{97.5} & \textbf{55.0} & \textbf{87.5} \\
\bottomrule

\end{tabular}
}
\end{table}
\else
\begin{table}[t]
\caption{\strut Shared RD head with and without reusing the original task-specific AVE on UniVTAC (SR, \%; $n=40$ trials per task).}
\label{tab:supp_shared}
\begin{tabular}{@{}c|cccccc|c@{}}
\toprule\noalign{\vskip 1.2pt}
\supphead{Setting} & \supphead{Insert\\Hole} & \supphead{Insert\\Tube} & \supphead{Lift\\Bottle} & \supphead{Lift\\Can} & \supphead{Pull Out\\Key} & \supphead{Put Bottle\\in Shelf} & \supphead{Avg.} \\
\midrule
\supphead{AVE off} & 95.0 & 87.5 & 95.0 & 80.0 & 92.5 & 52.5 & 83.8 \\
\supphead{Original AVE} & \textbf{97.5} & \textbf{92.5} & \textbf{100.0} & \textbf{82.5} & \textbf{97.5} & \textbf{55.0} & \textbf{87.5} \\
\bottomrule
\end{tabular}
\end{table}
\fi
\iflatexml
\begin{table}[t]
\renewcommand{\thetable}{E.1}
\renewcommand{\theHtable}{supp.thresholds}
\centering
\caption{\strut Offline validation-calibrated Exec AVE thresholds $(\tau_C, \tau_R)$ on UniVTAC for the adopted seed-42 models.}
\label{tab:supp_thresholds}
{\footnotesize
\begin{tabular}{lcccccc}
\toprule
\shortstack{Threshold} & \shortstack{Insert\\Hole} & \shortstack{Insert\\Tube} & \shortstack{Lift\\Bottle} & \shortstack{Lift\\Can} & \shortstack{Pull Out\\Key} & \shortstack{Put Bottle\\in Shelf} \\
\midrule
\shortstack{$\tau_C$} & 0.000000 & 0.005000 & 0.016075 & 0.005000 & 0.048692 & 0.010000 \\
\shortstack{$\tau_R$} & 0.163342 & 0.529858 & 0.571246 & 0.601553 & 0.193111 & 1.029039 \\
\bottomrule

\end{tabular}
}
\end{table}
\else
\begin{table}[t]
\renewcommand{\thetable}{E.1}
\renewcommand{\theHtable}{supp.thresholds}
\caption{\strut Offline validation-calibrated Exec AVE thresholds $(\tau_C, \tau_R)$ on UniVTAC for the adopted seed-42 models.}
\label{tab:supp_thresholds}
\begin{tabular}{@{}c|cccccc@{}}
\toprule\noalign{\vskip 1.2pt}
\supphead{Threshold} & \supphead{Insert\\Hole} & \supphead{Insert\\Tube} & \supphead{Lift\\Bottle} & \supphead{Lift\\Can} & \supphead{Pull Out\\Key} & \supphead{Put Bottle\\in Shelf} \\
\midrule
\supphead{$\tau_C$} & 0.000000 & 0.005000 & 0.016075 & 0.005000 & 0.048692 & 0.010000 \\
\supphead{$\tau_R$} & 0.163342 & 0.529858 & 0.571246 & 0.601553 & 0.193111 & 1.029039 \\
\bottomrule
\end{tabular}
\end{table}
\fi
\else
\iflatexml
\begin{table}[t]
\centering
\caption{\strut Shared RD head with and without reusing the original task-specific AVE on UniVTAC (SR, \%; $n=40$ trials per task).}
\label{tab:supp_shared}
{\footnotesize
\begin{tabular}{lccccccc}
\toprule
\shortstack{Setting} & \shortstack{Insert\\Hole} & \shortstack{Insert\\Tube} & \shortstack{Lift\\Bottle} & \shortstack{Lift\\Can} & \shortstack{Pull Out\\Key} & \shortstack{Put Bottle\\in Shelf} & \shortstack{Avg.} \\
\midrule
\shortstack{AVE off} & 95.0 & 87.5 & 95.0 & 80.0 & 92.5 & 52.5 & 83.8 \\
\shortstack{Original AVE} & \textbf{97.5} & \textbf{92.5} & \textbf{100.0} & \textbf{82.5} & \textbf{97.5} & \textbf{55.0} & \textbf{87.5} \\
\bottomrule

\end{tabular}
}
\end{table}

\begin{table}[t]
\renewcommand{\thetable}{E.1}
\renewcommand{\theHtable}{supp.thresholds}
\centering
\caption{\strut Offline validation-calibrated Exec AVE thresholds $(\tau_C, \tau_R)$ on UniVTAC for the adopted seed-42 models.}
\label{tab:supp_thresholds}
{\footnotesize
\begin{tabular}{lcccccc}
\toprule
\shortstack{Threshold} & \shortstack{Insert\\Hole} & \shortstack{Insert\\Tube} & \shortstack{Lift\\Bottle} & \shortstack{Lift\\Can} & \shortstack{Pull Out\\Key} & \shortstack{Put Bottle\\in Shelf} \\
\midrule
\shortstack{$\tau_C$} & 0.000000 & 0.005000 & 0.016075 & 0.005000 & 0.048692 & 0.010000 \\
\shortstack{$\tau_R$} & 0.163342 & 0.529858 & 0.571246 & 0.601553 & 0.193111 & 1.029039 \\
\bottomrule

\end{tabular}
}
\end{table}
\else
\begin{table*}[t]
\captionsetup{skip=3pt,justification=raggedright,singlelinecheck=false,strut=on,font={small,stretch=1},labelsep=period}
\noindent
\begin{minipage}[t]{\dimexpr(\textwidth-\columnsep)/2\relax}
\vspace{0pt}
\caption{\strut Shared RD head with and without reusing the original task-specific AVE on UniVTAC (SR, \%; $n=40$ trials per task).}
\label{tab:supp_shared}
\end{minipage}\hfill
\begin{minipage}[t]{\dimexpr(\textwidth-\columnsep)/2\relax}
\vspace{0pt}
\renewcommand{\thetable}{E.1}
\renewcommand{\theHtable}{supp.thresholds}
\caption{\strut Offline validation-calibrated Exec AVE thresholds $(\tau_C, \tau_R)$ on UniVTAC for the adopted seed-42 models.}
\label{tab:supp_thresholds}
\end{minipage}
\par\nointerlineskip
\noindent
\begin{minipage}[t]{\dimexpr(\textwidth-\columnsep)/2\relax}
\vspace{0pt}
\centering
\linespread{0.95}\footnotesize
\renewcommand{\arraystretch}{1.12}
\setlength{\aboverulesep}{1.3pt}\setlength{\belowrulesep}{1.3pt}
\setlength{\tabcolsep}{1.7pt}
\resizebox{\linewidth}{!}{%
\begin{tabular}{@{}c|cccccc|c@{}}
\toprule\noalign{\vskip 1.2pt}
\supphead{Setting} & \supphead{Insert\\Hole} & \supphead{Insert\\Tube} & \supphead{Lift\\Bottle} & \supphead{Lift\\Can} & \supphead{Pull Out\\Key} & \supphead{Put Bottle\\in Shelf} & \supphead{Avg.} \\
\midrule
\supphead{AVE off} & 95.0 & 87.5 & 95.0 & 80.0 & 92.5 & 52.5 & 83.8 \\
\supphead{Original AVE} & \textbf{97.5} & \textbf{92.5} & \textbf{100.0} & \textbf{82.5} & \textbf{97.5} & \textbf{55.0} & \textbf{87.5} \\
\bottomrule
\end{tabular}%
}
\end{minipage}\hfill
\begin{minipage}[t]{\dimexpr(\textwidth-\columnsep)/2\relax}
\vspace{0pt}
\centering
\linespread{0.95}\footnotesize
\renewcommand{\arraystretch}{1.12}
\setlength{\aboverulesep}{1.3pt}\setlength{\belowrulesep}{1.3pt}
\setlength{\tabcolsep}{1.7pt}
\resizebox{\linewidth}{!}{%
\begin{tabular}{@{}c|cccccc@{}}
\toprule\noalign{\vskip 1.2pt}
\supphead{Threshold} & \supphead{Insert\\Hole} & \supphead{Insert\\Tube} & \supphead{Lift\\Bottle} & \supphead{Lift\\Can} & \supphead{Pull Out\\Key} & \supphead{Put Bottle\\in Shelf} \\
\midrule
\supphead{$\tau_C$} & 0.000000 & 0.005000 & 0.016075 & 0.005000 & 0.048692 & 0.010000 \\
\supphead{$\tau_R$} & 0.163342 & 0.529858 & 0.571246 & 0.601553 & 0.193111 & 1.029039 \\
\bottomrule
\end{tabular}%
}
\end{minipage}
\par\vspace{4pt} 
\end{table*}
\fi
\fi

Reusing the original AVE without retraining raises shared-head success on every task, from 83.8\% to 87.5\%, or 201 to 210 of 240 trials in Table~\ref{tab:supp_shared}. The AVE-on row matches Configuration C in Table~\ref{tab:supp_timing_gain}.

\section{Implementation and Evaluation Details}
\label{sec:supp_details}
\setcounter{table}{1}

\subsection{Formulations, Calibration, and Execution}
\label{sec:supp_train_exec}
\allowdisplaybreaks[4] 

\paragraph{RD alignment and noise sharing.}
For reference chunk $A$ at $t_0$ with $H = 50$ and $N = 25$, each correction step $t \in \mathcal{T}_{t_0} = \{t_0+5, \dots, t_0+N\}$ with offset $g = t - t_0$ pairs $A_{g+j}$ with replanned chunk $B_j^{(t)}$ at absolute step $t + j = t_0 + g + j$ over $W = 5$ steps, keeping $g + W \le N + W \le H$. Target generation runs 10-step Euler flow matching on the frozen planner using the same initial noise sample for $A$ and $B^{(t)}$ along each trajectory, eliminating sampling variance. Both Smooth L1 inputs are normalized by per-dimension training scales. Planner training uses gradient accumulation of one.

\paragraph{Plan AVE formulation.}
At planning time $t_0$, given $V$ camera embeddings $\{v^{(i)}\}_{i=1}^{V}$ and $L = 4$ cached tactile embeddings $\{z^{(l)}\}_{l=1}^{L}$ sampled every $\Delta k = 5$ steps, the six reliability inputs in Eq.~(3) of the main paper are
\iflatexml
\begin{equation}
\begin{gathered}
v = \frac{1}{V}\sum_{i=1}^{V} v^{(i)}, \\
c_{\mathrm{mv}} = \frac{2}{V(V-1)}\sum_{1 \le i < j \le V} \frac{\langle v^{(i)}, v^{(j)} \rangle}{\|v^{(i)}\|_2 \|v^{(j)}\|_2 + \epsilon},
\end{gathered}
\label{eq:plan_feat1}
\end{equation}
\begin{equation}
\begin{gathered}
e_{\mathrm{track}} = \frac{1}{N_{\mathrm{prev}}}\sum_{\tau=1}^{N_{\mathrm{prev}}} \| q_\tau - a_\tau^{\mathrm{ref}} \|_2, \\
c_{\mathrm{temp}} = \frac{1}{L-1}\sum_{l=2}^{L} \frac{\langle z^{(l)}, z^{(l-1)} \rangle}{\|z^{(l)}\|_2 \|z^{(l-1)}\|_2 + \epsilon},
\end{gathered}
\label{eq:plan_feat2}
\end{equation}
\begin{equation}
\begin{gathered}
\delta_T = \| z_{t_0} - z_{t_0 - \Delta k} \|_2, \\
e_{\mathrm{pred}} = \frac{1}{2}\sum_{i \in \{L, R\}} v_{t_0}^{(i)} \| z_{t_0}^{(i)} - \hat{z}_{t_0}^{(i)} \|_2,
\end{gathered}
\label{eq:plan_feat3}
\end{equation}
\else
\begin{align}
&\begin{gathered}
v = \frac{1}{V}\sum_{i=1}^{V} v^{(i)}, \\
c_{\mathrm{mv}} = \frac{2}{V(V-1)}\sum_{1 \le i < j \le V} \frac{\langle v^{(i)}, v^{(j)} \rangle}{\|v^{(i)}\|_2 \|v^{(j)}\|_2 + \epsilon},
\end{gathered}\label{eq:plan_feat1}\\
&\begin{gathered}
e_{\mathrm{track}} = \frac{1}{N_{\mathrm{prev}}}\sum_{\tau=1}^{N_{\mathrm{prev}}} \| q_\tau - a_\tau^{\mathrm{ref}} \|_2, \\
c_{\mathrm{temp}} = \frac{1}{L-1}\sum_{l=2}^{L} \frac{\langle z^{(l)}, z^{(l-1)} \rangle}{\|z^{(l)}\|_2 \|z^{(l-1)}\|_2 + \epsilon},
\end{gathered}\label{eq:plan_feat2}\\
&\begin{gathered}
\delta_T = \| z_{t_0} - z_{t_0 - \Delta k} \|_2, \\
e_{\mathrm{pred}} = \frac{1}{2}\sum_{i \in \{L, R\}} v_{t_0}^{(i)} \| z_{t_0}^{(i)} - \hat{z}_{t_0}^{(i)} \|_2,
\end{gathered}\label{eq:plan_feat3}
\end{align}
\fi
where $c_{\mathrm{mv}} = 1$ for $V = 1$, $e_{\mathrm{track}} = 0$ at episode start and uses aligned end-effector poses under cross-embodiment transfer, and $v_{t_0}^{(i)} \in \{0, 1\}$ flags valid cached tactile predictions $\hat{z}_{t_0}^{(i)}$. Mapping the pooled tactile query $\bar{E}_Q$ to offsets $(\gamma_V, \gamma_T) = \mathrm{MLP}_q(\bar{E}_Q)$ yields the bounded attention weights in Eq.~(4) of the main paper:
\iflatexml
\begin{equation}
w_m = \mathrm{clip}\!\left(2\,\sigma\!\left(\log\frac{r_m+\epsilon}{1-r_m+\epsilon}+\gamma_m\right),w_{\min},w_{\max}\right),\quad m\in\{V,T\},
\label{eq:plan_weight}
\end{equation}
\else
\begin{equation}
\begin{gathered}
\begin{aligned}
w_m = \mathrm{clip}\!\Bigl(&2\,\sigma\!\left(\log\frac{r_m + \epsilon}{1 - r_m + \epsilon} + \gamma_m\right),\\
&w_{\min},\; w_{\max}\Bigr),
\end{aligned}\\
 m \in \{V, T\},
\label{eq:plan_weight}
\end{gathered}
\end{equation}
\fi
with $\epsilon = 10^{-6}$, $w_{\min} = 0.1$, and $w_{\max} = 2.0$.

\paragraph{Exec AVE supervision and decision rule.}
At each five-step check $t \in \mathcal{T}_{t_0}$ with $g = t - t_0$, the tactile-prediction error vector $u_t$ is
\iflatexml
\begin{equation}
u_t=\left[v_t^{(L)}\|z_t^{(L)}-\hat z_t^{(L)}\|_2,\;v_t^{(L)},\;\frac{t-t_{\mathrm{pred}}^{(L)}}{H},\;v_t^{(R)}\|z_t^{(R)}-\hat z_t^{(R)}\|_2,\;v_t^{(R)},\;\frac{t-t_{\mathrm{pred}}^{(R)}}{H}\right].
\label{eq:exec_ut}
\end{equation}
\else
\begin{equation}
\begin{aligned}
u_t = \Big[\,&v_t^{(L)}\|z_t^{(L)} - \hat{z}_t^{(L)}\|_2,\; v_t^{(L)},\; \frac{t - t_{\mathrm{pred}}^{(L)}}{H},\\
&v_t^{(R)}\|z_t^{(R)} - \hat{z}_t^{(R)}\|_2,\; v_t^{(R)},\; \frac{t - t_{\mathrm{pred}}^{(R)}}{H}\,\Big].
\end{aligned}
\label{eq:exec_ut}
\end{equation}
\fi
Five-step offline arm-action errors for KEEP, REPLAN, and CORRECT over the candidate grid $\mathcal{A} = \{0.5, 0.75, 1.0, 1.25, 1.5\}$ are
\iflatexml
\begin{equation}
\begin{gathered}
E_K = \frac{1}{5}\sum_{j=0}^{4} \| A_{g+j}^{\mathrm{arm}} - a_{t+j}^{\mathrm{demo,arm}} \|_2, \\
E_R = \frac{1}{5}\sum_{j=0}^{4} \| B_{j}^{(t),\mathrm{arm}} - a_{t+j}^{\mathrm{demo,arm}} \|_2,
\end{gathered}
\label{eq:exec_ek_er}
\end{equation}
\begin{equation}
\begin{gathered}
E_C(\alpha) = \frac{1}{5}\sum_{j=0}^{4} \| (A_{g+j}^{\mathrm{arm}} + \alpha \Delta \hat{a}_t) - a_{t+j}^{\mathrm{demo,arm}} \|_2, \\
\alpha_t^\star = \arg\min_{\alpha \in \mathcal{A}} E_C(\alpha), \\
E_C = E_C(\alpha_t^\star).
\end{gathered}
\label{eq:exec_ec}
\end{equation}
\else
\begin{align}
&\begin{gathered}
E_K = \frac{1}{5}\sum_{j=0}^{4} \| A_{g+j}^{\mathrm{arm}} - a_{t+j}^{\mathrm{demo,arm}} \|_2, \\
E_R = \frac{1}{5}\sum_{j=0}^{4} \| B_{j}^{(t),\mathrm{arm}} - a_{t+j}^{\mathrm{demo,arm}} \|_2,
\end{gathered}\label{eq:exec_ek_er}\\
&\begin{gathered}
E_C(\alpha) = \frac{1}{5}\sum_{j=0}^{4} \| (A_{g+j}^{\mathrm{arm}} + \alpha \Delta \hat{a}_t) - a_{t+j}^{\mathrm{demo,arm}} \|_2, \\
\alpha_t^\star = \arg\min_{\alpha \in \mathcal{A}} E_C(\alpha), \\
E_C = E_C(\alpha_t^\star).
\end{gathered}\label{eq:exec_ec}
\end{align}
\fi
Normalized error reductions relative to KEEP supervise predicted operation utilities $\hat{g}_m$, separate from the scaling gain $\alpha_t$:
\begin{equation}
\begin{gathered}
g_m^\star = \frac{E_K - E_m}{\max(E_K, \epsilon)}, \quad m \in \{C, R\}; \\
[\hat{g}_C, \hat{g}_R] = \mathrm{Linear}_2(h_t), \\
\alpha_t = 1.5\,\sigma(\mathrm{Linear}_1(h_t)).
\label{eq:op_benefit}
\end{gathered}
\end{equation}
In Eq.~(7) of the main paper, $\mathcal{L}_{\mathrm{gain}} = \sum_{m \in \{C, R\}} \mathrm{SmoothL1}(\hat{g}_m, g_m^\star)$, while $\mathrm{SmoothL1}_{\mathrm{valid}}(\alpha_t, \alpha_t^\star)$ is masked by $\mathbb{I}[\|\Delta \hat{a}_t\|_2 > 10^{-4}]$. At inference, $\mathrm{Gate}(h_t)$ selects
\begin{equation}
\begin{gathered}
s_K = 0, \\
s_C = \hat{g}_C - \tau_C, \\
s_R = \hat{g}_R - \tau_R, \\
m_t = \arg\max_{m \in \{K, C, R\}} s_m.
\label{eq:exec_gate}
\end{gathered}
\end{equation}

\paragraph{Offline threshold calibration.}
Per UniVTAC task, trajectories 0--31 train Exec AVE, 32--39 calibrate $(\tau_C, \tau_R)$, and 40--49 serve as held-out offline evaluation; closed-loop rollouts use disjoint scene seeds and evaluation feedback is explicitly disabled during threshold selection. Under Eq.~\eqref{eq:exec_gate}, $\tau_C$ and $\tau_R$ act as utility penalties rather than latency conversions. On trajectories 32--39, we sweep candidate thresholds---combining a fixed grid with $\hat{g}_C$ quantiles for $\tau_C$ and relative competing-score quantiles for $\tau_R$---and pick the candidate minimizing five-step demonstration action RMSE while keeping raw REPLAN proposals below 20\% and every single operation below 95\%. Table~\ref{tab:supp_thresholds} reports the resulting thresholds for the adopted seed-42 models. Because this objective targets offline action error rather than closed-loop success, Sec.~\ref{sec:supp_exec_ave} evaluates closed-loop impact directly through timing and gain ablations. At runtime, planning caches features and plan origin, new touch updates history, and early replanning resets the plan origin and relative features.

\subsection{Simulation and Cross-Embodiment Transfer}
\label{sec:supp_sim_transfer}

Main UniVTAC experiments run in Isaac Sim 5.1, separate from Sec.~\ref{sec:supp_full_attn} in Isaac Sim 4.5. Both UniVTAC, spaced every two 120~Hz physics steps, and OpenViTac operate at 60~Hz, so 50-step chunks span 0.833~s and 25-step executions span 0.417~s.

For zero-shot Allegro transfer, the thumb and index finger pinch while middle and ring fingers hold preset poses; an offline lookup table maps the scalar gripper opening to 16 hand joints, and aligning the pinch frame provides the arm IK target while thumb and index sensors supply the two tactile channels. Because the source policy is trained on a single-DoF parallel Panda gripper without Allegro target data, this pinch lookup preserves a consistent command interface while isolating adaptation to Allegro's hand kinematics and contact dynamics. RD and ResiP residuals apply in the 7-D source arm space before mapping. Across all six tasks in both Franka-to-UR5e and Panda-to-Allegro transfer, the planner, correction head, and AVE retain exact source weights and source calibrated thresholds $(\tau_C, \tau_R)$ without target fine-tuning or recalibration.

\subsection{Real-World Setup, Baselines, and Latency}
\label{sec:supp_realworld}

\paragraph{Hardware and tasks.}
The UMI platform~\cite{chi2024umi} mounts on a dual-arm Tianji Marvin M6-CCS robot with 7~DoF per arm and 696~mm reach, equipped with Xense XTac UMI G1 Leader/Follower grippers and dual-fingertip Xense GS-PS visuotactile sensors. Tactile streams run at 120~Hz and RGB at 30~Hz, synchronized to 50~Hz for training and execution.

\iflatexml
\begin{enumerate}[label=\textbullet]
\item \textbf{Cylinder grasping} (single-arm, 150 demos): grasp the cylinder and place it into the basket; typical failures include missed grasp, slip or drop, and missed placement.
\item \textbf{Potato-chip grasping} (single-arm, 150 demos): grasp the chip and return it intact into its box; typical failures include missed grasp, drop, fracture, and failed placement.
\item \textbf{Cap twisting} (dual-arm, 260 demos): twist off the bottle cap and set it down; typical failures include slippage, contact loss, and failed placement.
\item \textbf{Medication bagging} (dual-arm, 600 demos): grasp the medication, place it inside the bag, and seal the bag; typical failures include failed placement and incomplete sealing.
\end{enumerate}
\else
\begin{itemize}[leftmargin=*,itemsep=2pt,topsep=3pt]
\item \textbf{Cylinder grasping} (single-arm, 150 demos): grasp the cylinder and place it into the basket; typical failures include missed grasp, slip or drop, and missed placement.
\item \textbf{Potato-chip grasping} (single-arm, 150 demos): grasp the chip and return it intact into its box; typical failures include missed grasp, drop, fracture, and failed placement.
\item \textbf{Cap twisting} (dual-arm, 260 demos): twist off the bottle cap and set it down; typical failures include slippage, contact loss, and failed placement.
\item \textbf{Medication bagging} (dual-arm, 600 demos): grasp the medication, place it inside the bag, and seal the bag; typical failures include failed placement and incomplete sealing.
\end{itemize}
\fi

\paragraph{8-D residual extension for fragile grasping.}
On OpenViTac \textit{Grasp Chip} and real-world \textit{Grasp a Potato Chip}, both RD and ResiP extend from 7-D arm residuals to 8-D by appending continuous left-finger position in meters:
\iflatexml
\begin{enumerate}[label=\textbullet]
    \item \textbf{Targets:} per-window differences $B_j^{(t),:8} - A_{g+j}^{:8}$ average over five steps and use per-dimension normalized Smooth L1 loss.
    \item \textbf{Architectures:} expanding state and reference actions from 7-D to 8-D expands the condition vector from 15-D to 17-D; RD updates its state projection to $\mathrm{Linear}(17, 64)$ and head to $\mathrm{Linear}(64, 8)$, while ResiP expands its MLP input from 143-D to 145-D and output to 8-D.
    \item \textbf{Execution:} every five steps, $a^{\mathrm{exec}}_{:8} = a^{\mathrm{ref}}_{:8} + \alpha_t \Delta \hat{a}_t$ is applied across the window and the finger channel is clipped to $[0, 0.039]\text{ m}$, enabling continuous aperture regulation while Exec AVE suppresses post-contact disturbances.
\end{enumerate}
\else
\begin{itemize}[leftmargin=*,itemsep=2pt,topsep=3pt]
    \item \textbf{Targets:} per-window differences $B_j^{(t),:8} - A_{g+j}^{:8}$ average over five steps and use per-dimension normalized Smooth L1 loss.
    \item \textbf{Architectures:} expanding state and reference actions from 7-D to 8-D expands the condition vector from 15-D to 17-D; RD updates its state projection to $\mathrm{Linear}(17, 64)$ and head to $\mathrm{Linear}(64, 8)$, while ResiP expands its MLP input from 143-D to 145-D and output to 8-D.
    \item \textbf{Execution:} every five steps, $a^{\mathrm{exec}}_{:8} = a^{\mathrm{ref}}_{:8} + \alpha_t \Delta \hat{a}_t$ is applied across the window and the finger channel is clipped to $[0, 0.039]\text{ m}$, enabling continuous aperture regulation while Exec AVE suppresses post-contact disturbances.
\end{itemize}
\fi

\paragraph{Real-world baselines in Fig.~4 of the main paper.}
All five models are fine-tuned on the same UMI demonstrations and evaluated over 20 trials per task: visuotactile \textbf{Fast-WAM}~\cite{yuan2026fastwam}, visuotactile \textbf{Xiaomi XR-1}~\cite{xiaomi2026xr1}, vision-only \textbf{$\pi_{0.5}$}, offline \textbf{TSB}, and \textbf{ReTac}. To avoid unsafe physical PPO on hardware and fragile chips, real-world TSB keeps the ResiP MLP head and fixed execution at $\Delta k = 5$ and $\alpha = 1$, trained offline with the same RD objective as ReTac.

\paragraph{Latency breakdown.}
Table~3 of the main paper reports per-call round-trip latency. Across ReTac's stages, mean round-trip latency is 2.61~ms for correction alone given pre-encoded tactile features, 8.32~ms for tactile encoding plus correction as reported in Table~3 of the main paper, 12.73~ms when further including AVE decision-making, and 120.35~ms for main-policy planning excluding pre-computed tactile encoding and standalone AVE. Five-step intervals span 83.3~ms at 60~Hz and 100.0~ms at 50~Hz, both shorter than a 120.35~ms planning call.


\end{document}